\documentclass[journal]{IEEEtran}
\usepackage{amsmath,amsfonts}
\usepackage{algorithmic}
\usepackage{array}
\usepackage[caption=false,font=normalsize,labelfont=sf,textfont=sf]{subfig}
\usepackage{textcomp}
\usepackage{stfloats}
\usepackage{url}
\usepackage{hyperref}
\usepackage{verbatim}
\usepackage{graphicx}
\usepackage{cite}
\usepackage{multirow}
\usepackage{makecell}
\usepackage{booktabs}
\usepackage{tabularx}
\usepackage{amssymb}
\usepackage{ulem}
\usepackage{hhline}
\usepackage[table]{xcolor}
\usepackage[most]{tcolorbox}
\tcbuselibrary{breakable,listings}
\usepackage[ruled,linesnumbered]{algorithm2e}
\def\BibTeX{{\rm B\kern-.05em{\sc i\kern-.025em b}\kern-.08em
    T\kern-.1667em\lower.7ex\hbox{E}\kern-.125emX}}
\usepackage{balance}
\usepackage{upgreek}
\newtcblisting{rawbox}[1][]{
    enhanced, breakable,
    colback=gray!10, colframe=gray!50!black,
    boxrule=0.5pt,
    sharp corners,
    listing only,           
    notitle,
    #1
}

\begin{document}
\normalem
\title{Constructing Generalizable Parallel Algorithm Portfolios for Multi-Objective Binary Optimization via General-Purpose Instance Generation}

\author{Zhiyuan~Wang,~\IEEEmembership{Student Member,~IEEE},
Shengcai~Liu,~\IEEEmembership{Member,~IEEE}, Shaofeng~Zhang, 
and~Ke~Tang,~\IEEEmembership{Fellow,~IEEE}
\thanks{Zhiyuan Wang, Shengcai Liu, and Ke Tang are with the Guangdong Provincial Key Laboratory of Brain-inspired Intelligent Computation, Department of Computer Science and Engineering, Southern University of Science and Technology, Shenzhen 518055, China (email: wangzy2020@mail.sustech.edu.cn; liusc3@sustech.edu.cn; tangk3@sustech.edu.cn).

Shaofeng Zhang is with the Guangdong Provincial Key Laboratory of Brain-inspired Intelligent Computation, Department of Computer Science and Engineering, Southern University of Science and Technology, Shenzhen 518055, China, and Zhongguancun Academy, Beijing 100094, China (email: 12445025@mail.sustech.edu.cn).

Corresponding Author: Ke Tang
}
}


\markboth{IEEE TRANSACTIONS ON EVOLUTIONARY COMPUTATION,~Vol.~XX, No.~X, XX~XXXX}%
{Constructing Generalizable Parallel Algorithm Portfolios for Multi-Objective Binary Optimization via General-Purpose Instance Generation}

\maketitle

\begin{abstract}
    Despite progress in constructing generalizable parallel algorithm portfolios~(PAPs), no general-purpose method is yet available for multi-objective binary optimization problems~(MOBOPs) without problem-specific customization. To fill this gap, this work proposes domain-agnostic co-evolution of parameterized search for multi-objective binary optimization~(DACMO), the first general-purpose method for automatic PAP construction on MOBOPs. Its core, Unified Instance Representation~(UIR), represents and generates MOBOP instances with varying dimensions without knowledge of problem-specific instance structures. UIR separates common features of the target class from instance-specific features, preserving the common features learned from real instances during generation. DACMO co-evolves the training instances and PAP by alternating UIR-based instance augmentation with PAP updates. Experiments cover four structurally different MOBOP classes, including two black-box classes from real-world applications. Starting from five real training instances per class, DACMO constructs high-performance PAPs without problem-specific customization. On unseen instances from all four classes, these PAPs outperform portfolios of tuned classical multi-objective evolutionary algorithms and those constructed by the recent automatic PAP construction method EoH-S. DACMO also achieves overall performance comparable to the strong CEPS baseline and significantly outperforms it on two classes, without the substantial expert effort CEPS requires for problem-specific instance generator design.
\end{abstract}


\section{Introduction}
\label{sec:introduction}

\IEEEPARstart{M}{any} real-world applications, including resource allocation, manufacturing scheduling, and social network analysis, involve multi-objective binary optimization problems~(MOBOPs), which require optimizing multiple conflicting objectives over binary decision variables~\cite{wei_multi-objective_2022,he_2023_greedycooperative,zhang_2024_improvednsgaii,hong_efficient_2021}. Despite their shared use of binary decision variables, MOBOP classes can differ substantially in their objective functions, constraints, and instance structures. Consequently, search strategies and parameter settings that perform well on one MOBOP class may be less effective on another~\cite{huang_survey_2020,li_comparison_2019}. For a new MOBOP class, selecting, configuring, or designing suitable algorithms is therefore an important step toward obtaining high-quality solutions.

However, manually developing suitable algorithms for each new MOBOP class requires both domain knowledge and algorithm design expertise, often entailing substantial human effort. This process involves choosing or designing an algorithm and determining its parameter settings, which together constitute an algorithm configuration. Automatic algorithm configuration and design automate these decisions~\cite{huang_survey_2020,tang2024learn}. These methods evaluate candidate configurations on training instances from the target problem class to guide the search~\cite{hutter_sequential_2011,lopez-ibanez_irace_2016,ansotegui_model-based_2015}. The search can yield either a single configuration or a parallel algorithm portfolio~(PAP) comprising multiple configurations~\cite{liu_automatic_2019}.

Compared with a single configuration, a PAP can achieve better overall performance by exploiting the complementary strengths of its members across instances~\cite{huberman1997economics,gomes_algorithm_2001}. A PAP runs its member configurations independently and in parallel on the same instance, returning the best Pareto set among their outputs according to a specified performance indicator. This execution model naturally benefits from modern multi-core computing resources~\cite{hockney2019parallel}. The same PAP also provides a candidate pool for automatic algorithm selection~(AAS)~\cite{xu_hydra_2010,kadioglu_isac_2010}. AAS selects a single configuration for each instance, enabling deployment with limited parallel computing resources. The PAP's performance therefore defines the upper bound on selection performance~\cite{tang2024learn}. This work focuses on automatically constructing PAPs with complementary configurations for a given MOBOP class.

Generalization is a core requirement in automatic PAP construction: the resulting PAP should perform well on unseen instances from the target MOBOP class. For a new MOBOP class, however, training instances are often scarce and may not provide adequate coverage of that class~\cite{smith-miles_generating_2015}. Constructing a PAP using only these instances can lead to overtuning: the portfolio performs well on the training instances but poorly on unseen ones. To mitigate this issue, existing approaches augment the training set with instances that are difficult for the current PAP, guiding the search toward configurations that perform well on these instances and thus complement the current PAP~\cite{liu_generative_2022,wang_asp_2024,guo2025enhanced}. A representative framework is CEPS~\cite{tang_few-shots_2021}, which co-evolves the PAP and the training set by alternating between hard instance generation and PAP evolution to improve generalization.

However, reliance on instance generators customized for individual problem classes remains a key barrier to applying such co-evolutionary approaches to new MOBOP classes. Although MOBOPs all use binary decision variables, their instance representations can differ substantially. For example, instances of the multi-objective knapsack problem~(MKP) are described by item weights and profits~\cite{jaszkiewicz_performance_2002}, whereas instances of the multi-objective contamination control problem~(MCCP) involve contamination growth, control effectiveness, and costs at each stage of the production line~\cite{hu_contamination_2010}. These differences require generators to be customized according to the target problem definition and make direct reuse across problem classes difficult. For each new MOBOP class, designing a dedicated instance generator therefore still requires domain expertise and substantial human effort. This motivates a general-purpose method for instance representation and generation that can be directly applied to new MOBOP classes without requiring customized generators.

This work proposes the first general-purpose method for automatic PAP construction on MOBOPs, called domain-agnostic co-evolution of parameterized search for multi-objective binary optimization~(DACMO). The core of DACMO is Unified Instance Representation~(UIR). Rather than directly encoding native instance structures, UIR learns instance characteristics from binary solutions sampled from real instances and their corresponding objective vectors. It separates features common to instances of the target MOBOP class from instance-specific features. New instances are generated in UIR form by varying instance-specific features while preserving the learned common features. UIR represents and generates multi-objective binary instances with different numbers of decision variables, accommodating the dimensional variation common within a MOBOP class.

Building on UIR, DACMO alternates PAP updates with hard instance generation to co-evolve the PAP and training instances. During PAP updates, DACMO combines search operator generation using a large language model~(LLM) with automatic parameter configuration. This combination expands the search for complementary configurations beyond predefined algorithm templates. DACMO thus automatically constructs generalizable PAPs without requiring experts to design dedicated instance generators or tailor algorithms to the target MOBOP class.

The main contributions of this work are summarized below.
\begin{enumerate}
    \item The first general-purpose method for automatic PAP construction on MOBOPs, namely DACMO, is proposed. It requires no expert customization of instance generators or search algorithm templates for the target MOBOP class.
    \item UIR is proposed for representing and generating multi-objective binary instances with varying dimensions. It requires no knowledge of problem-specific instance structures. UIR provides a general-purpose instance augmentation method for PAP construction and other automated solver construction tasks on MOBOPs.
    \item Experiments validate DACMO's general applicability and effectiveness across four structurally different MOBOP classes, including two black-box problem classes from real-world applications. DACMO constructs a separate PAP for each class without problem-specific customization of the method. Its PAPs achieve better overall performance than PAPs composed of widely used multi-objective evolutionary algorithms~(MOEAs) with tuned parameters. They also achieve better overall performance than PAPs constructed by EoH-S, a recent method for automatic PAP construction. Without problem-specific customization, DACMO achieves overall performance comparable to CEPS, which relies on expert-designed instance generators. It also outperforms CEPS on two of the four classes. The constructed PAPs also provide high-quality candidate pools for AAS.
\end{enumerate}

The remainder of this article is organized as follows. Section~II formulates the problem of automatic PAP construction for MOBOPs and reviews related work. Section~III presents UIR for general-purpose instance representation and generation. Section~IV describes the DACMO method and its automatic PAP construction process. Section~V reports the experimental results and analysis. Section~VI concludes the article.

\section{PAP Construction for Multi-Objective Binary Optimization}
\label{sec:preliminaries}
\subsection{Notation and Problem Description}
\label{subsec:problem_description}

This work focuses on automatic PAP construction for unconstrained MOBOPs. Consider a target problem class, with \(\Omega\) denoting its instance space. Each instance \(s\in\Omega\) has \(d_s\) binary decision variables and a solution space \(\mathcal{X}_s=\{0,1\}^{d_s}\). The objective vector of a solution \(\mathbf{x}\in\mathcal{X}_s\) is denoted by \(\mathbf{y}=\mathbf{F}_s(\mathbf{x})\in\mathbb{R}^{m}\), where \(m\geq2\) is the number of objectives. Instances within the same MOBOP class follow a common problem definition but may differ in their specific objective functions and numbers of decision variables.

Let \(\Theta\) denote the algorithm configuration space. Each configuration \(\theta\in\Theta\) specifies the algorithm structure, search operators, and parameter settings. A PAP \(P=\{\theta_1,\ldots,\theta_K\}\subseteq\Theta\) comprises \(K\) distinct configurations. It runs its member configurations independently and in parallel on the same instance, returning the best Pareto set among their outputs according to a specified performance indicator. Let \(f(\theta,s)\) denote the performance of configuration \(\theta\) on instance \(s\) under a given run budget. This work uses hypervolume~(HV) to measure Pareto set quality, with larger values indicating better performance. With the same budget assigned to each member, the performance of \(P\) on \(s\) is defined as
\begin{equation}
    \label{eq:pap_perf}
    f(P,s)=\max_{\theta\in P}f(\theta,s).
\end{equation}
For AAS that selects a single configuration from \(P\), \(f(P,s)\) gives the upper bound on selection performance under the same run budget and performance indicator.

While \(f(P,s)\) measures performance on an individual instance, PAP construction aims to achieve good performance on unseen instances from the target MOBOP class. Let \(p(s)\) denote the instance distribution over \(\Omega\). The objective is to construct a PAP \(P\subseteq\Theta\) of prescribed size \(K\) that maximizes expected generalization performance:
\begin{equation}
    \label{eq:generalization_obj}
    \max_{P}\int_{s\in\Omega}f(P,s)\,p(s)\,\mathrm{d}s.
\end{equation}
To aggregate performance across instances with different objective scales, DACMO uses the normalized HV \(\bar f\) defined in~\eqref{eq:uir_normalized_performance} in place of \(f\) in this objective. In practice, the instance distribution \(p\) is unknown. PAP construction therefore relies on performance evaluated on a finite training set \(T\subseteq\Omega\) to guide the search. However, when training instances are scarce or provide inadequate coverage of the target class, training performance may not accurately reflect generalization performance~\cite{liu_performance_2020}. Constructing generalizable PAPs from limited training instances is therefore a key challenge in automatic PAP construction.

\subsection{Automatic PAP Construction}
\label{subsec:automatic_pap_construction}

With a sufficiently large and representative training set \(T\), established methods can construct PAPs with good generalization by searching for complementary configurations. Representative methods include GLOBAL, PARHYDRA, CLUSTERING~\cite{lindauer_automatic_2017}, and PCIT~\cite{liu_automatic_2019}. These methods primarily search within a predefined configuration space \(\Theta\), whose base algorithms and search operators must still be selected or designed by experts.

LLM-based algorithm generation extends automated search to the design of candidate algorithms~\cite{zhang2025llm,zheng_2025_montecarlo}. FunSearch~\cite{romera-paredes_mathematical_2024}, EoH~\cite{liu_evolution_2024}, and ReEvo~\cite{ye_2024_reevo} combine LLMs with evolutionary search to generate and refine candidate algorithms using performance feedback. For automatic PAP construction, EoH-S~\cite{liu2026eohs} targets a set of complementary heuristics. It uses each candidate algorithm's performance across instances to guide algorithm generation and selects members according to their combined performance. LLM-based design reduces the human effort required for algorithm development~\cite{yao_multi-objective_2025,ha_pareto-grid-guided_2026}. However, algorithm generation and member selection in PAP construction still rely on performance feedback from a given training set. When training instances are scarce or provide inadequate coverage, the resulting PAPs can still suffer from overtuning and poor generalization to unseen instances.

To mitigate this generalization problem, CEPS~\cite{tang_few-shots_2021} incorporates instance generation into PAP construction through the co-evolution of configuration and instance populations. After initialization, CEPS alternates between the following two stages.
\begin{enumerate}
    \item PAP evolution: With \(T\) fixed, search for a PAP \(P'\) to maximize \(\sum_{s\in T}f(P',s)\), then update \(P\leftarrow P'\).
    \item Training set evolution: With \(P\) fixed, generate an instance set \(T'\) to minimize \(\sum_{s\in T'}f(P,s)\), then update \(T\leftarrow T\cup T'\).
\end{enumerate}

However, CEPS relies on instance generation operators customized for the target problem class. Applying this mechanism directly to new MOBOP classes therefore requires a general-purpose method for instance representation and generation.

\subsection{Instance Representation and Generation}
\label{subsec:instance_generation_related_work}

Instance generation methods based on native instance structures typically require customization for the target problem class. For example, CEPS~\cite{tang_few-shots_2021} generates traveling salesman problem instances by perturbing city coordinates. Its vehicle routing application additionally requires tailored generation operators for customer pickup and delivery demands and service time windows. These operators depend on the respective problem definitions and are difficult to reuse directly across problem classes. Customizing these generators becomes even more difficult when solution evaluation relies on simulations or external programs and little information about native instance structures is available.

To improve PAP generalization, generated instances must preserve the common features of the target problem class while exposing weaknesses of the current PAP. To this end, DACE~\cite{wang2025evolving} learns neural instance representations from binary solutions sampled from real instances and their corresponding objective values. These representations enable instance generation without customized problem-specific generators. However, its representation and generation mechanisms are designed for single-objective binary optimization and cannot directly handle the multiple conflicting objectives in MOBOPs. Its network architecture also requires a fixed input dimension. The original study reports overfitting and difficulty achieving low validation error when training instances span a wider range of dimensions. These limitations prevent DACE from being directly applied to settings involving multiple objectives or varying instance dimensions. The following sections present DACMO, a general-purpose method for automatically constructing generalizable PAPs on MOBOPs without problem-specific customization.

\section{General-Purpose Instance Representation and Generation for Multi-Objective Binary Optimization}
\label{sec:uir}

UIR is the core of DACMO and provides a general-purpose method for instance representation and generation across MOBOP classes. Within the target MOBOP class, UIR represents and generates instances with varying numbers of decision variables. During generation, it preserves common features learned from real instances of that class, enabling instance augmentation without problem-specific generators.

\subsection{Architecture of UIR}
\label{subsec:uir_architecture}

A UIR is a neural representation of a MOBOP instance. It maps binary solutions to objective vectors. For each real instance of the target MOBOP class, binary solutions are sampled and evaluated to collect input--output pairs. Learning instance characteristics from these observations allows UIR to represent instances from any unconstrained MOBOP class without directly encoding native instance structures.

\begin{figure}[!t]
    \centering
    \includegraphics[width=0.8\linewidth]{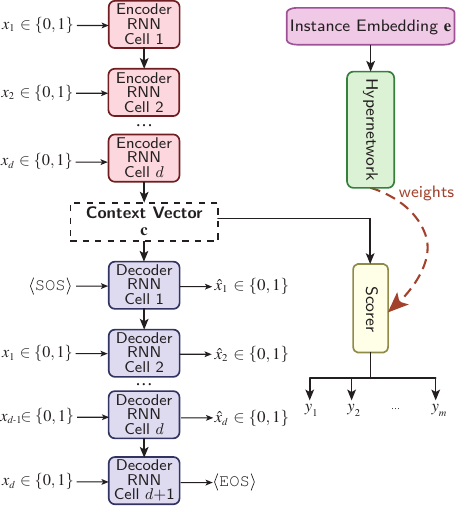}
    \caption{Architecture of UIR for multi-objective binary instances with varying dimensions. The Seq2Seq module encodes variable-length binary solutions into fixed-dimensional context vectors for the scorer. The parameters of the encoder, decoder, and hypernetwork are shared across instances of the target MOBOP class. The hypernetwork generates scorer parameters from each instance embedding. New instances are generated by varying the embedding while keeping the learned shared parameters fixed.}
    \label{fig:uir_architecture}
\end{figure}

Generating new instances of the target MOBOP class requires preserving its common features. UIR therefore separates features common to instances of the target MOBOP class from instance-specific features. As illustrated in Fig.~\ref{fig:uir_architecture}, the parameters of an encoder \(F_E\), a decoder \(F_D\), and a hypernetwork \(F_H\) are shared across instances of the target MOBOP class to capture common features. Instance-specific features are represented by a learnable embedding \(\mathbf{e}\) for each instance. The encoder \(F_E\) and decoder \(F_D\) together form a module for solution encoding. The hypernetwork \(F_H\) maps the instance embedding \(\mathbf{e}\) to the parameters of a scorer \(F_S\) for multi-objective evaluation. This design allows new instances to be generated by varying \(\mathbf{e}\) while keeping the learned shared parameters fixed.

Instances within the target MOBOP class may have different numbers of decision variables. The shared encoder \(F_E\) therefore uses a gated recurrent unit~(GRU) network to process binary solutions of varying lengths. For an input solution \(\mathbf{x}\), the final hidden states of all GRU layers are concatenated to form the fixed-dimensional context vector \(\mathbf{c}\in\mathbb{R}^{d_c}\). The scorer \(F_S\) then maps \(\mathbf{c}\) to the objective vector \(\mathbf{y}\in\mathbb{R}^{m}\):
\begin{equation}
    \label{eq:uir_evaluation}
    \begin{aligned}
        \mathbf{c} &= F_E(\mathbf{x};\mathbf{w}_E),\\
        \mathbf{y} &= F_S(\mathbf{c};\mathbf{w}_S),
    \end{aligned}
\end{equation}
where \(\mathbf{w}_E\) and \(\mathbf{w}_S\) denote the parameters of the encoder \(F_E\) and the scorer \(F_S\), respectively. This allows UIR to represent multi-objective instances of varying dimensions using the same encoder and scorer architectures. The scorer \(F_S\) is implemented as a multilayer perceptron~(MLP).

The parameters of the scorer \(F_S\) are generated from the instance embedding \(\mathbf{e}\in\mathbb{R}^{d_e}\) by the hypernetwork \(F_H\):
\begin{equation}
    \label{eq:uir_hypernetwork}
    \mathbf{w}_S=F_H(\mathbf{e};\mathbf{w}_H),
\end{equation}
where \(d_e\) is the embedding dimension and \(\mathbf{w}_H\) denotes the parameters of the hypernetwork \(F_H\). The hypernetwork is implemented as an MLP, and its parameters are shared across instances of the target MOBOP class.

To preserve structural features of the input binary solution \(\mathbf{x}\) in the context vector \(\mathbf{c}\), UIR uses an input reconstruction task during training. The decoder \(F_D\) also uses a GRU network. Its hidden states are initialized with the corresponding final encoder states to reconstruct the input solution as
\begin{equation}
    \label{eq:uir_reconstruction}
    \widehat{\mathbf{x}}=F_D(\mathbf{c};\mathbf{w}_D),
\end{equation}
where \(\widehat{\mathbf{x}}\) denotes the reconstructed solution and \(\mathbf{w}_D\) denotes the parameters of the decoder \(F_D\). Together, \(F_E\) and \(F_D\) form a sequence-to-sequence~(Seq2Seq) module. The shared parameters \(\mathbf{w}_E\), \(\mathbf{w}_D\), and \(\mathbf{w}_H\), together with the embeddings of individual instances, are learned jointly from samples collected from real instances. The training procedure is presented in Section~\ref{subsec:uir_training}.

\subsection{Learning UIRs from Real Instances}
\label{subsec:uir_training}

Given the training instance set \(T=\{s_1,\ldots,s_{|T|}\}\) of the target MOBOP class, binary solutions are sampled and evaluated on each instance \(s_i\) to construct a dataset
\[
    \mathcal{D}_i=\left\{\left(\mathbf{x}_k^i,\mathbf{y}_k^i\right)\right\}_{k=1}^{N_i},
\]
where \(N_i\) is the number of samples collected from \(s_i\), \(\mathbf{x}_k^i\in\mathcal{X}_{s_i}\) is a sampled solution, and \(\mathbf{y}_k^i\in\mathbb{R}^{m}\) is its true objective vector. Instances in \(T\) may have different numbers of decision variables \(d_{s_i}\). To fit the sampled solution--objective pairs, UIR jointly learns the shared parameters \(\mathbf{w}_E\), \(\mathbf{w}_D\), and \(\mathbf{w}_H\), together with an instance embedding \(\mathbf{e}_i\) for each instance \(s_i\). This yields a UIR for each real instance in \(T\).

To provide a reusable initialization across MOBOP classes, the Seq2Seq module is pretrained once on randomly generated binary sequences. Pretraining minimizes the cross-entropy loss for input reconstruction, encouraging context vectors to retain structural features of the binary sequences. This procedure requires neither objective values nor native structures of real instances.

For training on the target MOBOP class, the parameters \(\mathbf{w}_E\) and \(\mathbf{w}_D\) of the encoder and decoder are initialized from pretraining. The parameters \(\mathbf{w}_H\) of the hypernetwork and the instance embeddings \(\mathbf{e}_i\) are randomly initialized. The Seq2Seq module processes the sampled solutions as binary sequences of varying lengths. Start and end tokens mark sequence boundaries, while padding tokens align sequence lengths within each batch. For a sample \((\mathbf{x},\mathbf{y})\in\mathcal{D}_i\), let \(\mathbf{y}'_i\) denote the objective vector predicted by the UIR for \(s_i\) from \(\mathbf{x}\). The shared parameters and all instance embeddings are jointly learned by gradient descent with the following objective:
\begin{equation}
    \label{eq:uir_training_loss}
    \begin{aligned}
        \min_{\substack{\mathbf{w}_E,\mathbf{w}_D,\mathbf{w}_H,\\
        \mathbf{e}_1,\ldots,\mathbf{e}_{|T|}}}
        \sum_{i=1}^{|T|}\sum_{(\mathbf{x},\mathbf{y})\in\mathcal{D}_i}
        &\Bigl[\operatorname{MSE}\!\left(\mathbf{y},\mathbf{y}'_i\right)\\
        &\phantom{\Bigl[}{}-\frac{\lambda}{|T|\,d_{s_i}}
        \sum_{j=1}^{d_{s_i}}\mathbf{t}_j^{\top}\log\mathbf{p}_j
        \Bigr].
    \end{aligned}
\end{equation}
Here, \(\operatorname{MSE}\) denotes the mean squared error between the true objective vector \(\mathbf{y}\) and the predicted objective vector \(\mathbf{y}'_i\). The vector \(\mathbf{t}_j\) is the one-hot encoding of the value of the \(j\)th binary variable in the input solution, and \(\mathbf{p}_j\) is the probability distribution produced by the decoder at that position. The coefficient \(\lambda>0\) controls the relative weight of objective prediction and input reconstruction. Dividing the reconstruction loss by \(d_{s_i}\) yields the average error per position for sequences of different lengths. The factor \(1/|T|\) further scales this term inversely with the number of training instances.

The objective prediction term enables UIR to learn the multi-objective evaluation mappings of real instances, while the reconstruction term preserves structural features of binary solutions in the context vectors during training. Samples from multiple instances of the target MOBOP class jointly update the shared parameters, while samples from each instance also update its corresponding embedding. The shared parameters and instance embeddings thus capture common features within the target class and instance-specific features, respectively.

\subsection{Instance Generation with UIR}
\label{subsec:uir_generation}

UIR generates new MOBOP instances by varying instance embeddings. Its Seq2Seq module supports arbitrary numbers of decision variables in principle, allowing the dimension of each generated instance to be specified as needed. To preserve the common features of the target MOBOP class throughout generation, UIR keeps the learned parameters \(\mathbf{w}_E\), \(\mathbf{w}_D\), and \(\mathbf{w}_H\) fixed. UIR then perturbs an existing instance embedding \(\mathbf{e}\) to obtain a new embedding \(\mathbf{e}'\), from which the hypernetwork \(F_H\) generates new scorer parameters. The resulting solution-to-objective mapping defines a new MOBOP instance, which we refer to as a UIR instance and denote by \(\tilde{s}(\mathbf{e}')\).

In DACMO, performance feedback from the current PAP \(P\) guides the search for new instance embeddings. To compare performance across UIR instances with different objective scales, DACMO normalizes the attained HV against that of an instance-specific reference Pareto set. For each UIR instance \(\tilde{s}\), this set consists of the nondominated solutions among \(N_{\mathrm{ref}}\) randomly sampled and evaluated binary solutions. For maximization objectives, the HV reference point is the componentwise minimum of all objective vectors observed on \(\tilde{s}\) during random sampling and PAP execution. The HV of the reference set and that of each PAP member's output are computed using this same finalized reference point. The normalized PAP performance is defined as
\begin{equation}
    \label{eq:uir_normalized_performance}
    \bar f(P,\tilde{s})
    =\frac{f(P,\tilde{s})}{\operatorname{HV}_{\mathrm{ref}}(\tilde{s})},
\end{equation}
where \(\operatorname{HV}_{\mathrm{ref}}(\tilde{s})\) denotes the HV of the reference Pareto set. Consistent with the definition of PAP performance, \(f(P,\tilde{s})\) is the best HV attained by any member configuration on \(\tilde{s}\).

A lower normalized HV indicates a more challenging instance for the current PAP under the same run budget. With \(P\) fixed, the search for such instances is therefore formulated as
\begin{equation}
    \label{eq:uir_generation_objective}
    \min_{\mathbf{e}\in\mathbb{R}^{d_e}}
    \bar f\!\left(P,\tilde{s}(\mathbf{e})\right).
\end{equation}
Each evaluation of the search objective in~\eqref{eq:uir_generation_objective} requires running \(P\) on the corresponding UIR instance and computing its normalized performance. PGPE~\cite{sehnke_parameter-exploring_2010} is used for this continuous black-box search, and the UIR instance with the lowest normalized HV found during the search is retained. The PGPE procedure and its hyperparameter settings are provided in Appendix~A of the supplementary material.

Searching the instance embedding space provides new training instances without generation operators customized for the target MOBOP class. These instances guide subsequent configuration search toward improving PAP performance. The next section describes how DACMO integrates this generation mechanism into the co-evolution of the PAP and the training instance set.

\begin{algorithm}[!t]
    \small
    \caption{DACMO}
    \label{alg:dacmo}
    \SetKwInput{KwData}{Input}
    \SetKwInput{KwResult}{Output}
    \SetKw{To}{to}
    \SetKw{Break}{break}
    \SetKw{Return}{return}
    \SetCommentSty{itshape}
    \DontPrintSemicolon
    \KwData{Real training instances \(T\), PAP size \(K\), number \(n\) of candidate configurations per round, maximum number \(R_{\max}\) of co-evolution rounds.}
    \KwResult{The final PAP \(P\).}
    \tcp{Initialization}
    Learn UIRs for the instances in \(T\) using Section~\ref{subsec:uir_training}\nllabel{line:dacmo_learn_uir}\;
    \(M\leftarrow\) UIR instances defined by the learned UIRs\nllabel{line:dacmo_initialize_m}\;
    \(P\leftarrow\) initialize \(K\) members by cycling through NSGA-II, NSGA-III, MOEA/D, and SPEA2 with recommended parameter settings\nllabel{line:dacmo_initialize_p}\;
    Tune each member of \(P\) with SMAC3 in parallel to maximize \(\sum_{\tilde{s}\in M}\bar f(P,\tilde{s})\), keeping the other initialized members fixed\nllabel{line:dacmo_tune_initial_p}\;
    \For{\(r\leftarrow 1\) \To \(R_{\max}\)}{
        \tcp{PAP evolution}
        \(P_{\mathrm{cand}}\leftarrow P\)\nllabel{line:dacmo_initialize_candidates}\;
        \For{\(i\leftarrow 1\) \To \(n\)\nllabel{line:dacmo_candidate_loop_begin}}{
            Generate an algorithm and its configuration space \(\Theta_i'\) using an LLM with feedback from \(P\) on \(M\)\nllabel{line:dacmo_llm_generation}\;
            Use SMAC3 to find \(\theta_i'\in\Theta_i'\) maximizing \(\sum_{\tilde{s}\in M}\bar f(P\cup\{\theta_i'\},\tilde{s})\)\nllabel{line:dacmo_configure_candidate}\;
            \(P_{\mathrm{cand}}\leftarrow P_{\mathrm{cand}}\cup\{\theta_i'\}\)\nllabel{line:dacmo_add_candidate}\;
        }(\nllabel{line:dacmo_candidate_loop_end})
        \(P\leftarrow\displaystyle\operatorname*{arg\,max}_{\substack{P'\subseteq P_{\mathrm{cand}}\\|P'|=K}}\sum_{\tilde{s}\in M}\bar f(P',\tilde{s})\)\nllabel{line:dacmo_select_p}\;
        \tcp{Training set evolution}
        \lIf{\(r=R_{\max}\)}{\Break\nllabel{line:dacmo_final_round}}
        Reuse the normalized performance values \(\bar f(P,\tilde{s})\) obtained for all \(\tilde{s}\in M\) during PAP evolution\nllabel{line:dacmo_reuse_performance}\;
        \(M_{\mathrm{pool}}\leftarrow M,\quad M_{\mathrm{new}}\leftarrow\emptyset\)\nllabel{line:dacmo_initialize_pool}\;
        \For{\(j\leftarrow 1\) \To \(\lfloor|M|/2\rfloor+1\)\nllabel{line:dacmo_instance_loop_begin}}{
            Select \(\tilde{s}\) uniformly at random from \(M\)\;
            Generate \(\tilde{s}'\) by searching from the embedding of \(\tilde{s}\) as in Section~\ref{subsec:uir_generation}\nllabel{line:dacmo_generate_instance}\;
            \lIf{\(\bar f(P,\tilde{s}')\geq\max_{\tilde{s}\in M_{\mathrm{pool}}}\bar f(P,\tilde{s})\)}{\Break\nllabel{line:dacmo_stop_generation}}
            Select \(\tilde{s}^{\star}\) uniformly at random from instances in \(M_{\mathrm{pool}}\) with \(\bar f(P,\tilde{s}^{\star})>\bar f(P,\tilde{s}')\)\nllabel{line:dacmo_select_comparator}\;
            \(M_{\mathrm{pool}}\leftarrow M_{\mathrm{pool}}\setminus\{\tilde{s}^{\star}\}\)\;
            \(M_{\mathrm{new}}\leftarrow M_{\mathrm{new}}\cup\{\tilde{s}'\}\)\nllabel{line:dacmo_add_instance}\;
        }
        \(M\leftarrow M\cup M_{\mathrm{new}}\)\nllabel{line:dacmo_augment_m}\;
    }
    \Return{\(P\)}\nllabel{line:dacmo_return_p}\;
\end{algorithm}

\section{Domain-Agnostic Co-Evolution of Parameterized Search for Multi-Objective Binary Optimization~(DACMO)}
\label{sec:dacmo_framework}

\begin{figure*}[!t]
    \centering
    \includegraphics[width=0.75\textwidth]{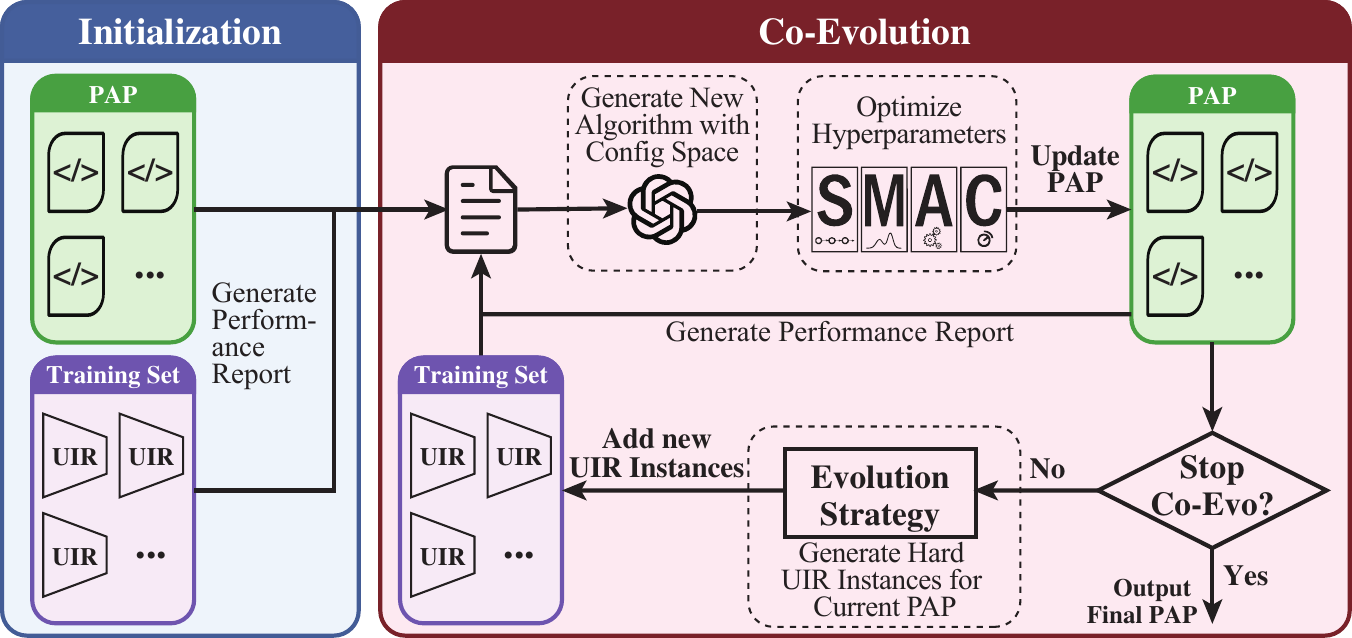}
    \caption{Overall workflow of DACMO. After initialization, DACMO alternates between PAP evolution and training set evolution. PAP evolution combines search operator generation using an LLM with automatic parameter configuration to identify complementary configurations. Training set evolution searches instance embeddings to generate UIR instances on which the current PAP attains low normalized HV. These instances guide subsequent PAP evolution to improve generalization.}
    \label{fig:dacmo}
\end{figure*}

DACMO builds on the co-evolutionary framework of CEPS~\cite{tang_few-shots_2021} to construct generalizable PAPs for a target MOBOP class. UIR provides a general-purpose instance generation mechanism for this process, eliminating the need for problem-specific generators. Given a limited set of real training instances, DACMO learns their UIRs and initializes a PAP. It then alternates between PAP evolution and training set evolution. PAP evolution combines LLM-based search operator generation with automatic parameter configuration to identify complementary configurations using performance feedback on the current training set. During training set evolution, performance feedback from the updated PAP guides the generation of new UIR instances that expose the PAP's weaknesses. These instances augment the training set and guide the next round of PAP evolution. The overall workflow and pseudocode are presented in Fig.~\ref{fig:dacmo} and Algorithm~\ref{alg:dacmo}, respectively. 

\subsection{Initialization}
\label{subsec:dacmo_initialization}

DACMO first learns UIRs jointly from the real training instances in $T$ using the procedure described in Section~\ref{subsec:uir_training}. The UIR instances defined by these learned UIRs form the initial training set $M$ (lines~\ref{line:dacmo_learn_uir}--\ref{line:dacmo_initialize_m} of Algorithm~\ref{alg:dacmo}). During co-evolution, algorithm configurations are evaluated on $M$, which is augmented with newly generated UIR instances during training set evolution.

PAP initialization starts from four classical MOEAs: NSGA-II~\cite{deb_fast_2002}, NSGA-III~\cite{deb_evolutionary_2014}, MOEA/D~\cite{zhang_moead_2007}, and SPEA2~\cite{zitzler_2001_spea2}. DACMO selects $K$ initial members by cycling through these algorithms. Each member starts with the recommended parameter settings of its algorithm (line~\ref{line:dacmo_initialize_p} of Algorithm~\ref{alg:dacmo}). SMAC3~\cite{hutter_sequential_2011} then tunes these parameters to improve the initial PAP's performance (line~\ref{line:dacmo_tune_initial_p}). Each tuning task keeps the other initialized members fixed and searches for parameter settings that maximize \(\sum_{\tilde{s}\in M}\bar f(P,\tilde{s})\), the aggregate normalized PAP performance on $M$. These tasks are performed independently and in parallel, and the resulting configurations form the initial PAP $P$.


\begin{algorithm}[!t]
    \small
    \caption{MOEA Template}
    \label{alg:moea_framework}
    \SetKwInput{KwData}{Input}
    \SetKwInput{KwResult}{Output}
    \SetKw{Return}{return}
    \SetCommentSty{itshape}
    \DontPrintSemicolon
    \KwData{MOBOP instance \(s\), algorithm parameter settings, evaluation budget.}
    \KwResult{The Pareto set found.}
    \tcp{\(Q\) and \(S\) denote the current and offspring populations}
    \tcp{\(V_Q\) and \(V_S\) store their corresponding objective vectors}
    \(\mathcal{A}\leftarrow\emptyset\)\nllabel{line:moea_initialize_archive}\;
    Provide an objective evaluation interface for \(s\) that records every evaluated solution and its objective vector in \(\mathcal{A}\)\nllabel{line:moea_evaluation}\;
    Apply the algorithm parameter settings\;
    \tcp{The LLM generates the three operators used below}
    \(Q,V_Q\leftarrow\)\texttt{initialization()}\nllabel{line:moea_initialization}\;
    \While{the evaluation budget is not exhausted}{
        \(S,V_S\leftarrow\)\texttt{offspring\_generation(\(Q,V_Q\))}\nllabel{line:moea_offspring}\;
        \(Q,V_Q\leftarrow\)\texttt{update(\(Q,S,V_Q,V_S\))}\nllabel{line:moea_update}\;
    }
    \Return{the nondominated solutions recorded in \(\mathcal{A}\)}\nllabel{line:moea_return}\;
\end{algorithm}

\subsection{PAP Evolution}
\label{subsec:pap_evolution}

PAP evolution combines LLM-based search operator generation with automatic parameter configuration~\cite{van_stein_loop_nodate} to identify complementary configurations on the current training set \(M\). Each round comprises \(n\) candidate generation tasks based on the current PAP \(P\) (lines~\ref{line:dacmo_candidate_loop_begin}--\ref{line:dacmo_candidate_loop_end} of Algorithm~\ref{alg:dacmo}). For the \(i\)th task, DACMO summarizes each member's performance across the instances in \(M\) in a textual report. The LLM receives this report together with the members' algorithm implementations and parameter settings. These inputs guide the generation of a new MOEA and its configuration space \(\Theta_i'\) in line~\ref{line:dacmo_llm_generation}.

The MOEA template in Algorithm~\ref{alg:moea_framework} provides a common interface for configuring and evaluating the generated algorithms. The LLM designs and implements the population initialization, offspring generation, and population update operators in lines~\ref{line:moea_initialization}, \ref{line:moea_offspring}, and~\ref{line:moea_update}, respectively. It also specifies the tunable parameters and their ranges, thereby defining the corresponding configuration space \(\Theta_i'\). Generated code undergoes syntax and interface checks, followed by execution tests on small instances. Code that fails validation is regenerated by the LLM before parameter configuration. The LLM prompt templates and the accompanying MOEA code template are provided in Appendix~B of the supplementary material.

For each validated algorithm, SMAC3~\cite{hutter_sequential_2011} searches for a configuration \(\theta_i'\in\Theta_i'\) to maximize \(\sum_{\tilde{s}\in M}\bar f(P\cup\{\theta_i'\},\tilde{s})\) (line~\ref{line:dacmo_configure_candidate} of Algorithm~\ref{alg:dacmo}). Because PAP performance is determined by the best member on each instance, this objective evaluates how the candidate improves the combined performance across \(M\). Parameter configuration thus directly targets improvements in portfolio performance. The current \(P\) and \(M\) remain fixed throughout candidate generation and configuration, allowing the \(n\) tasks to run independently in parallel.

The original PAP members and the newly generated configurations form the candidate pool \(P_{\mathrm{cand}}\) (lines~\ref{line:dacmo_initialize_candidates} and~\ref{line:dacmo_add_candidate} of Algorithm~\ref{alg:dacmo}). To compare candidate portfolios on a consistent basis, DACMO uses a common HV reference point for all configurations on each training instance. For maximization objectives, this point is the componentwise minimum of all objective vectors observed during reference sampling and evaluation of the configurations in \(P_{\mathrm{cand}}\). Using this point, DACMO recomputes the HV values of the recorded configuration outputs and the reused reference Pareto set without additional solver runs. DACMO then enumerates all subsets containing \(K\) configurations and computes their normalized PAP performance on each instance as in~\eqref{eq:uir_normalized_performance}. The subset with the highest aggregate normalized PAP performance on \(M\) is selected as the updated \(P\) (line~\ref{line:dacmo_select_p}). This joint selection exploits the configurations' complementary strengths across training instances.

\subsection{Training Set Evolution}
\label{subsec:training_set_evo}

Training set evolution augments \(M\) with UIR instances that are more challenging for the updated PAP \(P\), providing new training feedback for subsequent PAP evolution. Under the same run budget, lower values of the normalized HV \(\bar f(P,\tilde{s})\) defined in~\eqref{eq:uir_normalized_performance} indicate more challenging instances for \(P\). With \(P\) fixed, DACMO reuses the normalized performance values obtained for \(P\) on \(M\) during PAP evolution to select new instances for augmentation. It initializes a comparison pool \(M_{\mathrm{pool}}\) with \(M\) and creates an empty set \(M_{\mathrm{new}}\) to store new instances selected for augmentation (lines~\ref{line:dacmo_reuse_performance}--\ref{line:dacmo_initialize_pool} of Algorithm~\ref{alg:dacmo}).

DACMO performs up to \(\lfloor |M|/2\rfloor+1\) instance generation iterations per round. In each iteration, an instance \(\tilde{s}\) is sampled uniformly at random from \(M\). The search described in Section~\ref{subsec:uir_generation} starts from this instance's embedding and generates a new UIR instance \(\tilde{s}'\) with the same number of decision variables as \(\tilde{s}\) (lines~\ref{line:dacmo_instance_loop_begin}--\ref{line:dacmo_generate_instance}). The new instance \(\tilde{s}'\) is selected for augmentation if its normalized HV is lower than that of at least one instance in \(M_{\mathrm{pool}}\). If this condition is not met, instance generation stops for the current round (line~\ref{line:dacmo_stop_generation}). Otherwise, DACMO selects \(\tilde{s}^{\star}\) uniformly at random from the instances in \(M_{\mathrm{pool}}\) with normalized HV greater than \(\bar f(P,\tilde{s}')\). DACMO removes \(\tilde{s}^{\star}\) from \(M_{\mathrm{pool}}\) and adds \(\tilde{s}'\) to \(M_{\mathrm{new}}\) (lines~\ref{line:dacmo_select_comparator}--\ref{line:dacmo_add_instance}).

After generation, DACMO augments \(M\) with \(M_{\mathrm{new}}\), retaining all existing training instances (line~\ref{line:dacmo_augment_m}). Performance feedback on the augmented \(M\) guides algorithm generation, parameter configuration, and PAP selection in the next round. In round \(R_{\max}\), DACMO returns \(P\) immediately after the PAP update (lines~\ref{line:dacmo_final_round} and~\ref{line:dacmo_return_p}).

\section{Computational Studies}
\label{sec:experiments}

This section evaluates DACMO on four structurally different MOBOP classes, including two black-box problem classes from real-world applications. For each class, PAP construction starts from a small set of real training instances, and the resulting PAPs are evaluated on unseen real instances from that class. The experiments address the following four research questions~(RQs):

\begin{description}
    \item[\textit{RQ1}:] Can DACMO be applied directly to different MOBOP classes without problem-specific customization and produce PAPs with better generalization performance than those constructed by existing automatic PAP construction methods?
    \item[\textit{RQ2}:] Can UIR generate instances of varying dimensions that preserve the common features of the target MOBOP class, and can these instances improve PAP generalization through instance augmentation?
    \item[\textit{RQ3}:] How do co-evolution and automatic algorithm design improve PAP generalization?
    \item[\textit{RQ4}:] Can the PAPs constructed by DACMO serve as high-quality candidate pools for AAS?
\end{description}

The four problem classes and their constraint-handling procedures are introduced next, followed by the compared methods and the experimental protocol. The subsequent subsections present the results and analyses addressing these RQs.

\subsection{Problem Instances and Constraint Handling}
\label{subsec:problem_instances}

For each problem class, PAP construction starts from five real training instances with 32, 40, 48, 56, and 64 decision variables, respectively. The resulting PAPs are evaluated on held-out real instances, with 30 independent instances at each of four dimensions: 32, 48, 64, and 100. The 100-dimensional instances assess generalization beyond the training dimension range. For MKP, MCCP, and MCIMP, testing also introduces changes in the instance generation parameters or underlying networks, as detailed below.

All four problem classes are formulated as maximization problems. For an instance \(s\), \(F_{s,j}(\mathbf{x})\) denotes the \(j\)th component of \(\mathbf{F}_s(\mathbf{x})\), and \(d_s\) is the number of decision variables. Infeasible solutions to the multi-objective knapsack problem~(MKP) and the multi-objective complementary influence maximization problem~(MCIMP) are repaired before objective evaluation. By incorporating repair into objective evaluation, both problems are formulated over the full binary search space without explicit constraints.

\subsubsection{Multi-Objective Match Max Problem~(MMMP)}
 MMMP generalizes OneMinMax~\cite{doerr_runtime_2016} by maximizing the number of matching bits between a binary solution and each of \(m\) reference vectors \(\mathbf{r}_j\in\{0,1\}^{d_s}\):
\begin{equation}
\label{eq:mmmp_objectives}
F_{s,j}(\mathbf{x})=d_s-D_{\mathrm H}(\mathbf{x},\mathbf{r}_j),
\qquad j=1,\ldots,m,
\end{equation}
where \(D_{\mathrm H}\) denotes Hamming distance. The problem is unconstrained. The experiments use \(m=3\), with reference vectors sampled subject to \(D_{\mathrm H}(\mathbf{r}_a,\mathbf{r}_b)>d_s/m\) for every pair \(a\ne b\).

\subsubsection{Multi-Objective Knapsack Problem~(MKP)}
 The bi-objective MKP~\cite{jaszkiewicz_performance_2002} maximizes two measures of total profit subject to a knapsack capacity constraint. An instance contains \(d_s\) items, each with weight \(w_i\) and profits \(v_{i1}\) and \(v_{i2}\). With \(x_i=1\) indicating selection of item \(i\), the objectives and constraint are
\begin{equation}
\label{eq:mkp_objectives}
\begin{gathered}
F_{s,j}(\mathbf{x})=\sum_{i=1}^{d_s}v_{ij}x_i,\qquad j=1,2,\\
\sum_{i=1}^{d_s}w_ix_i\leq w_{\max}.
\end{gathered}
\end{equation}
For an infeasible solution, selected items are scanned in index order. Once their cumulative weight exceeds \(w_{\max}\), the corresponding decision variable and all subsequent variables are set to zero.

Profits are sampled within \([0,1]\), using \(p\) and \(1-p\) as the split points for the first and second profit attributes, respectively. For each attribute, half of the items receive profits sampled from the lower subinterval and half from the upper subinterval. Sampling enforces \(w_a>w_b\Rightarrow v_{aj}>v_{bj}\) for \(j=1,2\), so heavier items have higher profits under both objectives. The parameter \(p\) is sampled from \([0.6,0.75]\) for training instances and \([0.65,0.8]\) for test instances.

\subsubsection{Multi-Objective Contamination Control Problem~(MCCP)}
 MCCP~\cite{hu_contamination_2010} models contamination control along a production line with \(d_s\) stages. It maximizes the sum of the probabilities that contamination is below a threshold at each stage while minimizing total control cost. At stage \(i\), \(x_i=1\) applies a control measure at cost \(c_i\). Given the contamination growth rate \(\alpha_i\) and reduction rate \(\Gamma_i\), the contamination level evolves as \(z_i=\alpha_i(1-x_i)(1-z_{i-1})+(1-\Gamma_i x_i)z_{i-1}.\)

The control costs \(c_i\) are sampled uniformly from \([0,1]\). The initial contamination level \(z_0\) and rates \(\Gamma_i\) and \(\alpha_i\) follow \(\operatorname{Beta}(1,\beta_{z_0})\), \(\operatorname{Beta}(1,\beta_\Gamma)\), and \(\operatorname{Beta}(1,\beta_\alpha)\), respectively. For training instances, \(\beta_{z_0}\), \(\beta_\Gamma\), and \(\beta_\alpha\) are sampled uniformly from \([20,28]\), \([4,6]\), and \([4,6]\), respectively. The corresponding ranges for test instances are \([24,32]\), \([5,7]\), and \([5,7]\). The contamination threshold \(u\) is set to \(0.1\) for training instances and \(0.15\) for test instances.

Objective evaluation uses \(N_{\mathrm{MC}}=1000\) Monte Carlo samples:
\begin{equation}
\label{eq:mccp_objectives}
\begin{gathered}
F_{s,1}(\mathbf{x})=\sum_{i=1}^{d_s}\frac{1}{N_{\mathrm{MC}}}
\sum_{\ell=1}^{N_{\mathrm{MC}}}\mathbb{I}\!\left(z_i^{(\ell)}<u\right),\\
F_{s,2}(\mathbf{x})=-\sum_{i=1}^{d_s}c_ix_i,
\end{gathered}
\end{equation}
where \(\mathbb{I}(\cdot)\) is the indicator function and \(z_i^{(\ell)}\) is the contamination level at stage \(i\) in sample \(\ell\). This bi-objective formulation imposes no additional constraints.

\subsubsection{Multi-Objective Complementary Influence Maximization Problem~(MCIMP)}
 MCIMP~\cite{hong_efficient_2021} jointly maximizes the spread of opinion \(B\) in a social network and minimizes seed cost. An instance comprises a social network \(\mathcal{G}=(V,E,p)\), a candidate seed set \(N\subseteq V\) of size \(d_s\), and an initial set \(S_A\) of nodes holding opinion \(A\). Here, \(p\) denotes the propagation probability. The interaction between opinions is specified by \(\mathbf{q}=[q_{A|\emptyset},q_{A|B},q_{B|\emptyset},q_{B|A}]^\top.\)
Setting \(x_i=1\) selects candidate node \(i\) as a seed for opinion \(B\). Its cost is \(c_i=\lambda_i r_i\), where \(r_i\) is the node degree and \(\lambda_i\in[0.8,1.2]\) is sampled independently for each node. The objectives and constraint are
\begin{equation}
\label{eq:mcimp_objectives}
\begin{gathered}
F_{s,1}(\mathbf{x})=\operatorname{ActiveNum}(\mathcal{G},N,S_A,\mathbf{q},\mathbf{x}),\\
F_{s,2}(\mathbf{x})=-\sum_{i=1}^{d_s}c_ix_i,\\
\sum_{i=1}^{d_s}x_i\leq k,
\end{gathered}
\end{equation}
where \(\operatorname{ActiveNum}\) estimates the number of nodes holding opinion \(B\) after propagation using 1000 Monte Carlo samples~\cite{lu_competition_2015}, and \(k\) is the maximum number of seeds. Infeasible solutions are repaired by retaining the first \(k\) selected nodes in index order and clearing all remaining selections.

Training instances use the Wiki network. For test instances, the network is sampled uniformly from Wiki, Facebook, and Epinions. The interaction vector \(\mathbf{q}\) is chosen with equal probability from \([0.5,0.75,0.5,0.75]^\top\) and \([0.5,0.25,0.5,0.25]^\top\)~\cite{lu_competition_2015}.

\subsection{Compared Methods}
\label{subsec:compared_methods}

Under the training and test settings described in Section~\ref{subsec:problem_instances}, the PAPs constructed by DACMO are first compared with portfolios of classical MOEAs:
\begin{itemize}
    \item \textbf{Default:} This portfolio consists of NSGA-II~\cite{deb_fast_2002}, NSGA-III~\cite{deb_evolutionary_2014}, MOEA/D~\cite{zhang_moead_2007}, and SPEA2~\cite{zitzler_2001_spea2}, each using the default parameter settings recommended in the literature.
    \item \textbf{Tuning:} This portfolio consists of the same four algorithms with parameters tuned by SMAC3~\cite{hutter_sequential_2011} on the initial real training set.
\end{itemize}

The comparison also includes \textbf{EoH-S}~\cite{liu2026eohs}, a recent LLM-based method for automatic algorithm portfolio design. EoH-S constructs its PAP using the fixed initial set of real training instances. For a fair comparison, every new algorithm generated by EoH-S is tuned using SMAC3. The tuning budget per algorithm is identical to that used in line~\ref{line:dacmo_configure_candidate} of Algorithm~\ref{alg:dacmo}. In our MOBOP implementation, each tuned configuration is evaluated using its normalized HV on every training instance. These performance values guide subsequent algorithm evolution and the selection of complementary PAP members.

\textbf{CEPS}~\cite{tang_few-shots_2021} incorporates instance augmentation into PAP construction through co-evolution but requires experts to customize an instance generator for each target MOBOP class. In our experiments, its generators follow the training instance generation mechanisms described in Section~\ref{subsec:problem_instances}. Its PAP update uses the same combination of LLM-based search operator generation and automatic parameter configuration as DACMO. This comparison assesses whether DACMO can construct PAPs without any problem-specific customization while achieving generalization performance comparable to that of CEPS PAPs.

Two ablation baselines further examine the contributions of UIR-based instance augmentation and automatic algorithm design:
\begin{itemize}
    \item \textbf{NoInsGen:} This ablation retains DACMO's LLM-based search operator generation and automatic parameter configuration. It constructs a PAP on the initial real training set without instance augmentation.
    \item \textbf{NoAlgDesign:} This ablation uses the full augmented training set from DACMO's final co-evolution round. It constructs a PAP by tuning the four classical MOEAs with SMAC3. This baseline assesses the additional benefit of automatic algorithm design over tuning fixed algorithm templates when augmented training data are available.
\end{itemize}

The PAP constructed by DACMO for each problem class is also evaluated as a candidate configuration pool for AAS, which selects a single configuration to solve each instance. This work focuses on PAP construction rather than AAS selector design. To address RQ4, three selection scenarios are therefore used to characterize candidate-pool quality independently of a specific selector. Oracle and Worst give the upper and lower performance bounds by taking the maximum and minimum member performance on each test instance. Random gives the expected performance under uniform random selection, calculated as the mean member performance. Practical AAS methods can often substantially outperform uniform random selection by exploiting instance information. Random therefore provides a baseline for assessing candidate-pool quality without relying on an informed selection strategy. The performance under these scenarios is compared with that of standalone NSGA-II, NSGA-III, MOEA/D, and SPEA2. The standalone baselines also include two more recent algorithms: AGE-MOEA-II~\cite{panichella_improved_2022} and DCEA~\cite{zheng_novel_2024}.

\subsection{Experimental Protocol}
\label{subsec:experimental_protocol}

Each PAP construction method produces one PAP of \(K=4\) member configurations for each target MOBOP class. DACMO and CEPS differ only in their instance generation mechanisms and otherwise use identical construction procedures and parameter settings. Both perform up to \(R_{\max}=4\) co-evolution rounds, generating and configuring \(n=20\) new algorithms per round. Each instance generation search is limited to 200 iterations. EoH-S maintains a population of four algorithms and generates up to 20 new algorithms per generation. Its total budget of 80 algorithm generation attempts matches that of DACMO and CEPS. All three methods use Kimi-K2.5 for algorithm generation. Across methods, the SMAC3 budget is fixed at 1600 trials per tuning task for both initial members and newly generated algorithms. During PAP construction, the number of function evaluations (\(\#\mathrm{FEs}\)) is limited to 1600 per configuration on each training instance.

DACMO learns UIRs from the five real training instances of each target MOBOP class using the procedure in Section~\ref{subsec:uir_training}. The sampled solution--objective pairs from each instance are split into training and validation sets. The sample size per instance, network architecture, and training hyperparameters are identical across the four problem classes, as summarized in Table~\ref{tab:uir_param}.

\begin{table}[t]
    \centering
    \caption{Data Sampling, Network Architecture, and Training Settings for UIR}
    \label{tab:uir_param}
    \footnotesize
    \setlength{\tabcolsep}{4pt}
    \renewcommand{\arraystretch}{1.12}
    \begin{tabularx}{\linewidth}{@{}>{\raggedright\arraybackslash}p{0.5\linewidth}>{\raggedright\arraybackslash}X@{}}
        \toprule
        \textbf{Parameter} & \textbf{Value} \\
        \midrule
        Samples per real training instance, \(N_i\) & \(100{,}000\) \\
        Training--validation split & \(3:1\) \\
        \midrule
        Token embedding dimension & 64 \\
        GRU hidden layer widths & \([128,128]\) \\
        Context vector dimension, \(d_c\) & 256 \\
        Instance embedding dimension, \(d_e\) & 32 \\
        Scorer MLP layer widths & \([128,128,m]\), \(m\) is \#objectives. \\
        Hypernetwork MLP layer widths & \([64,49408+129m]\) \\
        Hypernetwork activation & LeakyReLU \\
        \midrule
        Learning rate & Cosine annealing from 2e-3 to 5e-4 \\
        Batch size, Training epochs & 1024, 5000 \\
        Loss weight, \(\lambda\) & 1 \\
        \bottomrule
    \end{tabularx}
\end{table}

The final PAPs are evaluated on the same real test instances described in Section~\ref{subsec:problem_instances}. Each PAP member is assigned a budget of \(\#\mathrm{FEs}=1600\) per test instance. The standalone MOEAs use the same budget. All six standalone MOEAs use default parameter settings, with the four classical algorithms using the configurations in Default. The output of each run is the nondominated set extracted from all solutions evaluated during that run~\cite{tanabe_benchmarking_2017}.

To aggregate performance across instances with different objective scales, we use the normalized HV defined in~\eqref{eq:uir_normalized_performance}. For each instance, a reference Pareto set is constructed by randomly sampling and evaluating binary solutions and retaining the nondominated subset. The HV of this set provides the normalization denominator. These random solutions are sampled and evaluated independently of solver runs, and their evaluations do not count toward the solver's FE budget. During PAP construction, DACMO and CEPS normalize performance feedback using reference sets built from \(N_{\mathrm{ref}}=5000\) random solutions per training instance. For final testing, the reported performance is normalized using reference sets built from \(100{,}000\) random solutions per unseen real instance.

To ensure comparable HV values on each test instance, all methods within a comparison group use a common final reference point. Once the runs are complete, the objective ranges recorded by these methods are combined with those obtained from the random solutions. Under the maximization convention, the reference point is the componentwise minimum of the lower bounds of these ranges. The HVs of the algorithm output and the reference set are computed using this point, and their ratio gives the normalized score. This score can exceed 1 because the reference set is based on finite random samples.

Normalized HV is summarized by its mean and population standard deviation across test instances. For each problem class, these statistics are computed separately over the 30 instances at each dimension and over all 120 instances combined. Within each comparison group, all method pairs are compared using two-sided Wilcoxon signed-rank tests, with scores paired by test instance. The tests are conducted separately for each problem class, both at each dimension and across all four dimensions combined. Holm correction is applied to the full set of pairwise \(p\)-values within each analysis, with a significance level of \(0.05\).

All experiments are conducted using Python 3.10 on Ubuntu 22.04, with AMD EPYC CPUs and NVIDIA GPUs based on the Ampere and Ada Lovelace architectures. The source code and benchmark instances have been anonymously released at \url{https://github.com/AnonymousSubmitBot/DACMO}.

\subsection{Generalization Performance of Constructed PAPs}
\label{subsec:pap_generalization}

\begin{table*}[tbp]
    \centering
    \caption{Normalized HV (Mean $\pm$ Std.) of PAPs on unseen real instances. The highest mean before display rounding is \textbf{bolded} with a \colorbox{gray!70}{gray} background. \underline{Underlining} indicates no significant difference from the result with the highest mean under two-sided Wilcoxon tests with Holm correction at $\alpha=0.05$. Each dimension includes 30 instances, and Total aggregates all 120 instances of a class.}
    \label{tab:pap_perf}
    {
    \setlength{\aboverulesep}{0.05pt}
    \setlength{\belowrulesep}{0.05pt}
    \renewcommand{\arraystretch}{1.1}
    \begin{tabular}{cccccccc}
        \toprule
        Problem & Dim & DACMO & NoInsGen & NoAlgDesign & EoH-S & Tuning & Default \\
        \midrule
        \multirow{5}[1]{*}{MMMP} & 32 & \underline{1.289±0.047} & \cellcolor{gray!70}\textbf{1.294±0.056} & 1.269±0.054 & 1.251±0.049 & 1.225±0.050 & 1.175±0.054 \\
        \cmidrule{2-8}
         & 48 & \cellcolor{gray!70}\textbf{1.572±0.109} & 1.555±0.123 & 1.537±0.098 & 1.493±0.094 & 1.480±0.104 & 1.375±0.098 \\
        \cmidrule{2-8}
         & 64 & \cellcolor{gray!70}\textbf{1.827±0.102} & 1.746±0.114 & 1.754±0.102 & 1.683±0.078 & 1.661±0.106 & 1.494±0.085 \\
        \cmidrule{2-8}
         & 100 & \cellcolor{gray!70}\textbf{2.172±0.178} & 2.037±0.144 & 2.047±0.178 & 1.956±0.131 & 1.896±0.151 & 1.625±0.109 \\
        \cmidrule{2-8}
         & Total & \cellcolor{gray!70}\textbf{1.715±0.346} & 1.658±0.294 & 1.652±0.309 & 1.596±0.274 & 1.566±0.269 & 1.417±0.188 \\
        \midrule
        \multirow{5}[1]{*}{MKP} & 32 & \underline{0.990±0.025} & \underline{0.984±0.022} & \underline{0.992±0.028} & \cellcolor{gray!70}\textbf{0.994±0.024} & \underline{0.990±0.018} & \underline{0.983±0.017} \\
        \cmidrule{2-8}
         & 48 & \cellcolor{gray!70}\textbf{1.006±0.032} & \underline{0.990±0.024} & \underline{1.005±0.028} & \underline{1.003±0.028} & \underline{0.997±0.020} & 0.991±0.024 \\
        \cmidrule{2-8}
         & 64 & \cellcolor{gray!70}\textbf{1.010±0.029} & \underline{1.005±0.026} & \underline{1.005±0.026} & \underline{1.007±0.020} & 0.993±0.015 & 0.998±0.020 \\
        \cmidrule{2-8}
         & 100 & \underline{1.015±0.025} & 1.003±0.021 & \cellcolor{gray!70}\textbf{1.018±0.023} & \underline{1.012±0.024} & 0.995±0.022 & 0.999±0.019 \\
        \cmidrule{2-8}
         & Total & \cellcolor{gray!70}\textbf{1.006±0.029} & 0.995±0.025 & \underline{1.005±0.028} & \underline{1.004±0.025} & 0.994±0.019 & 0.992±0.021 \\
        \midrule
        \multirow{5}[1]{*}{MCCP} & 32 & \cellcolor{gray!70}\textbf{1.320±0.042} & 1.310±0.043 & 1.281±0.045 & 1.270±0.042 & 1.242±0.036 & 1.160±0.038 \\
        \cmidrule{2-8}
         & 48 & \cellcolor{gray!70}\textbf{1.682±0.095} & 1.599±0.090 & 1.531±0.077 & 1.511±0.085 & 1.410±0.071 & 1.272±0.062 \\
        \cmidrule{2-8}
         & 64 & \cellcolor{gray!70}\textbf{2.008±0.147} & 1.843±0.130 & 1.677±0.115 & 1.710±0.127 & 1.496±0.087 & 1.260±0.085 \\
        \cmidrule{2-8}
         & 100 & \cellcolor{gray!70}\textbf{2.352±0.210} & 2.181±0.218 & 1.889±0.135 & 1.830±0.149 & 1.572±0.116 & 1.274±0.083 \\
        \cmidrule{2-8}
         & Total & \cellcolor{gray!70}\textbf{1.840±0.407} & 1.734±0.348 & 1.595±0.243 & 1.581±0.239 & 1.430±0.148 & 1.242±0.084 \\
        \midrule
        \multirow{5}[1]{*}{MCIMP} & 32 & \cellcolor{gray!70}\textbf{1.164±0.126} & \underline{1.162±0.125} & 1.157±0.124 & 1.158±0.125 & 1.107±0.113 & 1.047±0.090 \\
        \cmidrule{2-8}
         & 48 & \cellcolor{gray!70}\textbf{1.350±0.205} & 1.346±0.204 & 1.244±0.154 & 1.340±0.206 & 1.191±0.131 & 1.088±0.114 \\
        \cmidrule{2-8}
         & 64 & \cellcolor{gray!70}\textbf{1.543±0.291} & 1.539±0.291 & 1.299±0.220 & 1.523±0.281 & 1.261±0.183 & 1.129±0.091 \\
        \cmidrule{2-8}
         & 100 & \cellcolor{gray!70}\textbf{1.827±0.480} & \underline{1.822±0.483} & 1.317±0.267 & 1.795±0.462 & 1.334±0.292 & 1.108±0.126 \\
        \cmidrule{2-8}
         & Total & \cellcolor{gray!70}\textbf{1.471±0.392} & 1.467±0.392 & 1.254±0.209 & 1.454±0.378 & 1.223±0.210 & 1.093±0.111 \\
        \bottomrule
    \end{tabular}
    }
\end{table*}

To address \textit{RQ1}, we evaluate whether DACMO can construct PAPs with better generalization performance for different MOBOP classes without problem-specific customization. Table~\ref{tab:pap_perf} compares the PAPs from DACMO, Default, Tuning, EoH-S, and the two ablation variants on unseen real instances. It reports the mean and standard deviation of normalized HV and the statistical test results. For each problem class, results are grouped by test dimension, and the \textit{Total} row summarizes all 120 test instances. The NoInsGen and NoAlgDesign results are analyzed under \textit{RQ2} and \textit{RQ3}, respectively.

DACMO uses the same construction procedure for all four MOBOP classes. The resulting PAPs achieve higher aggregate mean normalized HV than the Default, Tuning, and EoH-S PAPs in every class. The differences from Default and Tuning are statistically significant in every class. The gains over Tuning are particularly pronounced on MCCP and MCIMP, the two black-box problem classes from real-world applications. DACMO improves mean normalized HV by 28.7\% and 20.3\% on these classes, respectively. With the same algorithm generation and parameter tuning budgets as EoH-S, DACMO achieves corresponding gains of 16.4\% and 1.2\%, with both differences being statistically significant. On MMMP, the gains over Tuning and EoH-S are 9.5\% and 7.5\%, respectively, and both differences are also statistically significant. These results demonstrate that DACMO can automatically construct effective PAPs from small sets of real instances across structurally different benchmark and application problems.

DACMO also improves on the classical MOEA portfolios on MKP. Its gains over Default and Tuning are 1.3\% and 1.2\%, respectively. The corresponding gains for EoH-S are 1.2\% and 1.0\%. DACMO attains a higher aggregate mean than EoH-S (1.006 versus 1.004), but the difference is not statistically significant. A plausible explanation for these similar gains is that established MOEA search strategies are already well suited to MKP, a classical multi-objective optimization problem. DACMO's significant gains over Default and Tuning show that its general-purpose construction procedure also benefits this established benchmark.

The constructed PAPs also generalize beyond the dimensions used during construction. All instances used for PAP construction have at most 64 decision variables. On the 100-dimensional real test instances, DACMO achieves higher mean normalized HV than Default, Tuning, and EoH-S in all four classes. Its relative gains over EoH-S increase with test dimension on MMMP and MCCP, reaching 11.0\% and 28.5\%, respectively, at 100 dimensions. A single PAP per class therefore covers the tested dimensions both within and beyond the training range, without requiring a separate construction run for larger instances.

\begin{table}[tbp]
    \centering
    \caption{Normalized HV (mean $\pm$ standard deviation) of PAPs from DACMO and CEPS. The CEPS generators are customized by experts for each problem class. Bold, gray shading, and underlining follow Table~\ref{tab:pap_perf}.}
    \label{tab:ceps_perf}
    {
    \setlength{\aboverulesep}{0.05pt}
    \setlength{\belowrulesep}{0.05pt}
    \renewcommand{\arraystretch}{1.1}
    \setlength{\tabcolsep}{12pt}
    \begin{tabular}{cccc}
        \toprule
        Problem & Dim & DACMO & CEPS \\
        \midrule
        \multirow{5}[1]{*}{MMMP} & 32 & \underline{1.289±0.047} & \cellcolor{gray!70}\textbf{1.295±0.055} \\
        \cmidrule{2-4}
         & 48 & \cellcolor{gray!70}\textbf{1.572±0.109} & 1.551±0.105 \\
        \cmidrule{2-4}
         & 64 & \cellcolor{gray!70}\textbf{1.827±0.102} & 1.776±0.115 \\
        \cmidrule{2-4}
         & 100 & \cellcolor{gray!70}\textbf{2.172±0.178} & 2.085±0.174 \\
        \cmidrule{2-4}
         & Total & \cellcolor{gray!70}\textbf{1.715±0.346} & 1.677±0.314 \\
        \midrule
        \multirow{5}[1]{*}{MKP} & 32 & 0.990±0.025 & \cellcolor{gray!70}\textbf{1.016±0.047} \\
        \cmidrule{2-4}
         & 48 & \underline{1.006±0.032} & \cellcolor{gray!70}\textbf{1.018±0.038} \\
        \cmidrule{2-4}
         & 64 & \cellcolor{gray!70}\textbf{1.010±0.029} & \underline{1.009±0.023} \\
        \cmidrule{2-4}
         & 100 & \underline{1.015±0.025} & \cellcolor{gray!70}\textbf{1.017±0.017} \\
        \cmidrule{2-4}
         & Total & 1.006±0.029 & \cellcolor{gray!70}\textbf{1.015±0.033} \\
        \midrule
        \multirow{5}[1]{*}{MCCP} & 32 & \underline{1.320±0.042} & \cellcolor{gray!70}\textbf{1.320±0.041} \\
        \cmidrule{2-4}
         & 48 & 1.682±0.095 & \cellcolor{gray!70}\textbf{1.719±0.088} \\
        \cmidrule{2-4}
         & 64 & 2.008±0.147 & \cellcolor{gray!70}\textbf{2.122±0.165} \\
        \cmidrule{2-4}
         & 100 & 2.352±0.210 & \cellcolor{gray!70}\textbf{2.910±0.299} \\
        \cmidrule{2-4}
         & Total & 1.840±0.407 & \cellcolor{gray!70}\textbf{2.018±0.614} \\
        \midrule
        \multirow{5}[1]{*}{MCIMP} & 32 & \cellcolor{gray!70}\textbf{1.164±0.126} & 1.161±0.125 \\
        \cmidrule{2-4}
         & 48 & \cellcolor{gray!70}\textbf{1.350±0.205} & 1.346±0.202 \\
        \cmidrule{2-4}
         & 64 & \cellcolor{gray!70}\textbf{1.543±0.291} & 1.535±0.285 \\
        \cmidrule{2-4}
         & 100 & \cellcolor{gray!70}\textbf{1.827±0.480} & 1.811±0.472 \\
        \cmidrule{2-4}
         & Total & \cellcolor{gray!70}\textbf{1.471±0.392} & 1.463±0.384 \\
        \bottomrule
    \end{tabular}
    }
\end{table}

DACMO also constructs competitive PAPs without requiring experts to develop an instance generator for each target MOBOP class. Table~\ref{tab:ceps_perf} compares its PAPs with those constructed by CEPS using expert-designed generators. The two methods share the same co-evolutionary framework, PAP update procedure, and construction budgets. In this experiment, the CEPS generators are customized for each class according to the mechanisms used to generate its real training instances. CEPS therefore benefits from explicit knowledge of how instances are generated. By learning UIRs from five real training instances per class, DACMO can generate instances without requiring access to those mechanisms.

DACMO exceeds CEPS in aggregate mean normalized HV by 2.3\% on MMMP and 0.5\% on MCIMP, and both differences are statistically significant. On MKP and MCCP, its aggregate means reach 99.1\% and 91.2\% of the corresponding CEPS means, respectively. DACMO thus constructs competitive PAPs across MOBOP classes without the expert customization of instance generators required by CEPS.

\subsection{Effectiveness of UIR-Based Instance Generation}
\label{subsec:uir_generation_effectiveness}

To address \textit{RQ2}, we first examine whether UIR generates new instances that preserve the common features of the target MOBOP class. We then assess whether its instance generation mechanism improves PAP generalization. The visual analysis focuses on the three bi-objective classes: MKP, MCCP, and MCIMP. Shape similarity is more difficult to assess from static three-dimensional plots of the three-objective MMMP. For MMMP, we therefore quantify instance similarity through objective function matching.

\begin{figure*}[!t]
    \centering
    \subfloat[]{\includegraphics[width=0.325\textwidth]{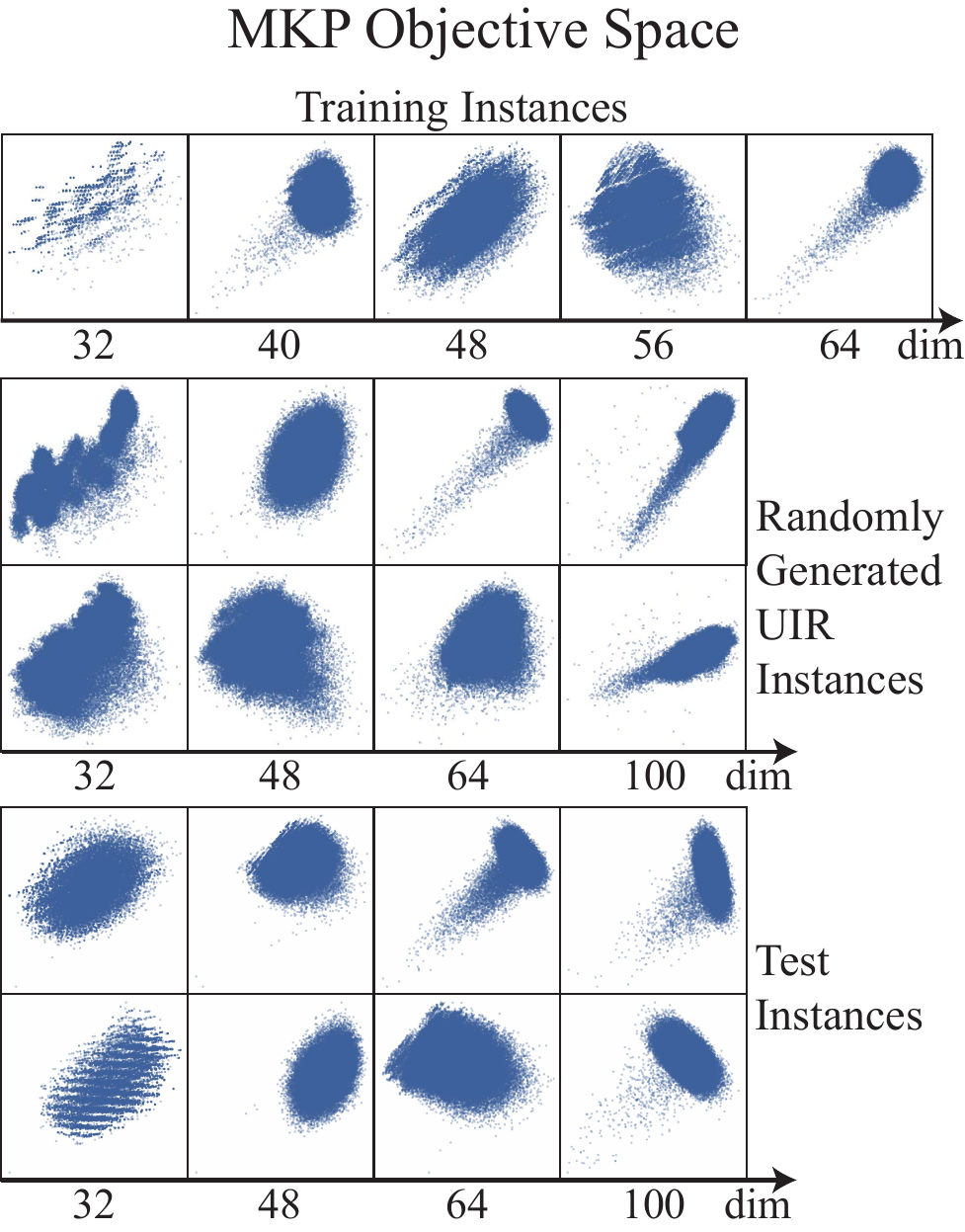}\label{fig:uir_mkp_vis}}
    \hfill
    \subfloat[]{\includegraphics[width=0.325\textwidth]{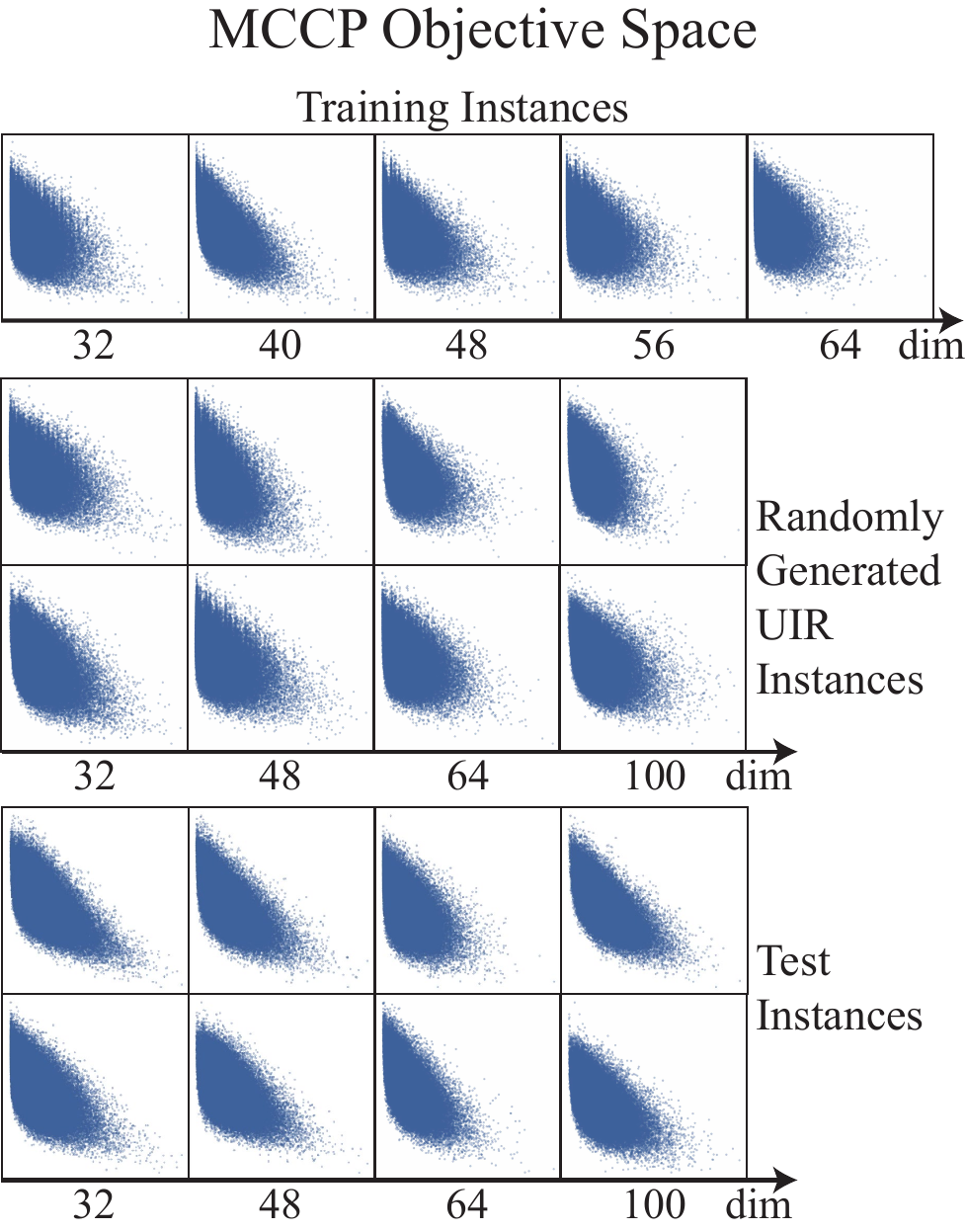}\label{fig:uir_mccp_vis}}
    \hfill
    \subfloat[]{\includegraphics[width=0.325\textwidth]{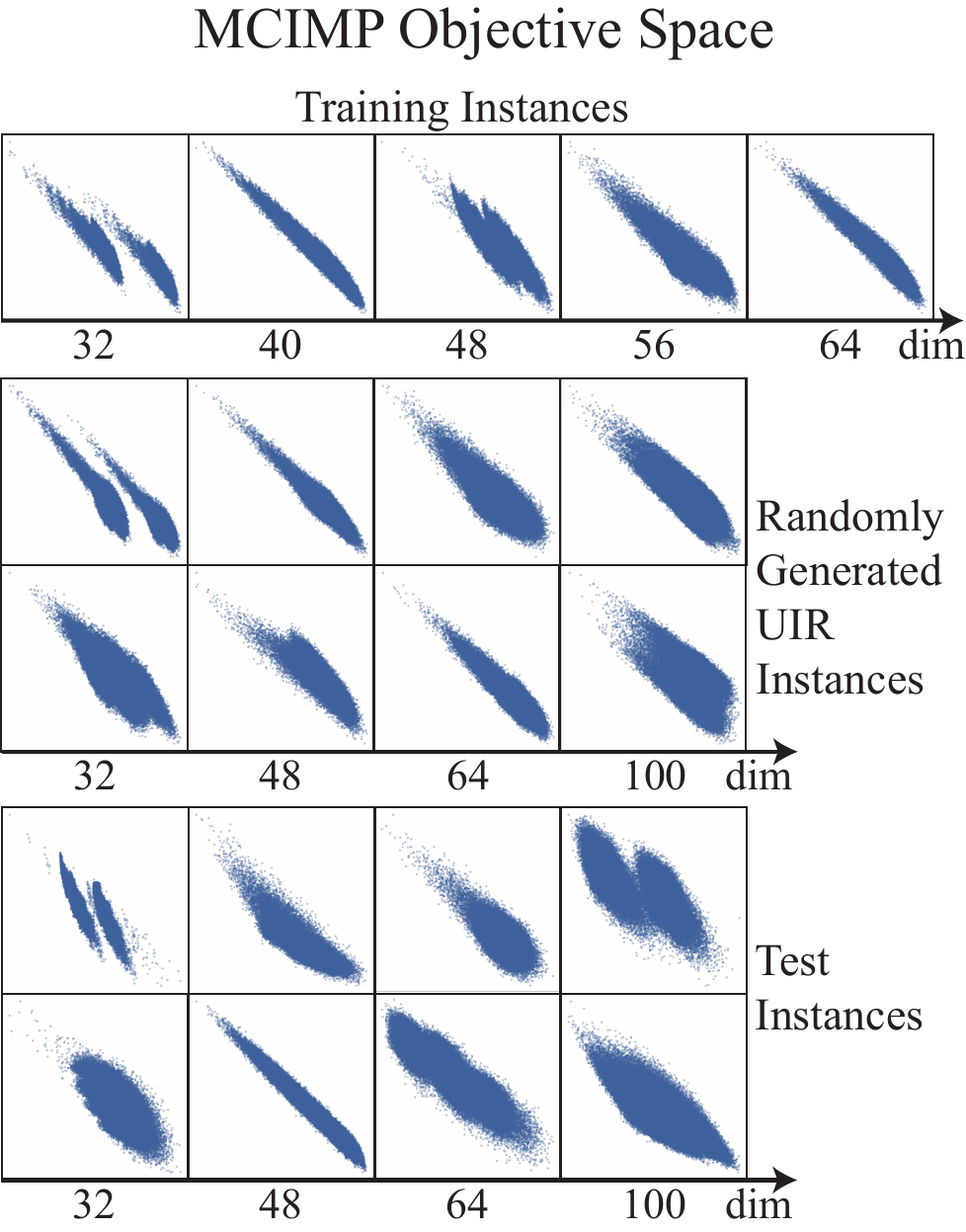}\label{fig:uir_mcimp_vis}}
    \caption{Objective-space distributions of real training instances (top), UIR instances generated using the trained model (middle), and real test instances (bottom). Each small panel plots the objective vectors of \(100{,}000\) randomly sampled binary solutions. The dimension labels indicate the numbers of decision variables. For each class, all five training instances and two generated and two test instances at each displayed dimension are shown. (a) MKP. (b) MCCP. (c) MCIMP.}
    \label{fig:uir_instance_vis}
\end{figure*}

Fig.~\ref{fig:uir_instance_vis} compares the objective-space distributions of real training instances, generated UIR instances, and real test instances. For this visualization, new instance embeddings are sampled from a normal distribution while the learned Seq2Seq and hypernetwork parameters remain fixed. For each real or generated instance, \(100{,}000\) binary solutions are randomly sampled and evaluated, and their objective vectors are plotted. Each problem class is represented by all five training instances, together with two generated instances and two test instances at each dimension of 32, 48, 64, and 100.

The generated UIR instances exhibit characteristic objective-space patterns of their target classes. These include clustered distributions on MKP, fan-shaped distributions with curved boundaries on MCCP, and elongated diagonal bands on MCIMP. Although the training instances have at most 64 decision variables, the generated 100-dimensional instances retain the characteristic distribution patterns of their target classes. The same representation and generation procedure captures these distinct patterns across MOBOP classes without problem-specific generator design.

The MMMP experiment compares the trained UIR model with a randomly initialized model of the same architecture. This comparison assesses the effect of training on the similarity between generated and real MMMP objective functions. Each model generates 32 instances at each dimension of 32, 48, 64, and 100. Instance embeddings are sampled from a standard normal distribution while the model parameters remain fixed.

For each generated UIR instance, BRKGA searches for the three binary reference vectors defining a real MMMP instance. The search minimizes the MSE between the objective values of the real and UIR instances on a given set of binary solutions. Before computing the MSE, objective values are min--max normalized to \([0,1]\) separately for each objective and instance. The search set contains \(10{,}000\) randomly sampled binary solutions, and both groups use the same search settings and budget. Validation MSE is then computed on a separate set of \(300{,}000\) random binary solutions, with normalization recomputed on this set.

\begin{table}[tbp]
    \centering
    \caption{Validation MSE (Mean $\pm$ Std.) between generated and matched real MMMP instances. The two UIR models share the same architecture and each generate 32 instances per dimension. MSE reduction denotes the percentage decrease in mean MSE relative to the randomly initialized model.}
    \label{tab:mmmp_similarity}
    {
    \footnotesize
    \setlength{\aboverulesep}{0.05pt}
    \setlength{\belowrulesep}{0.05pt}
    \renewcommand{\arraystretch}{1.1}
    \setlength{\tabcolsep}{3pt}
    \begin{tabular*}{\columnwidth}{@{\extracolsep{\fill}}cccc@{}}
        \toprule
        Dim & Randomly initialized & Trained on MMMP & MSE reduction \\
        \midrule
        32 & \(\text{6.2e-2}\pm\text{1.3e-2}\) & \(\text{5.5e-3}\pm\text{9.4e-4}\) & 91.1\% \\
        48 & \(\text{6.0e-2}\pm\text{1.3e-2}\) & \(\text{5.3e-3}\pm\text{7.2e-4}\) & 91.1\% \\
        64 & \(\text{7.3e-2}\pm\text{1.3e-2}\) & \(\text{5.4e-3}\pm\text{6.4e-4}\) & 92.6\% \\
        100 & \(\text{6.7e-2}\pm\text{1.4e-2}\) & \(\text{1.2e-2}\pm\text{6.0e-4}\) & 82.4\% \\
        \bottomrule
    \end{tabular*}
    }
\end{table}

Table~\ref{tab:mmmp_similarity} shows that instances generated by the trained UIR closely resemble real MMMP instances in their objective functions. The randomly initialized model produces substantially larger matching errors. At the training dimensions of 32, 48, and 64, the trained UIR achieves mean validation MSEs of only \(5.3\times10^{-3}\)--\(5.5\times10^{-3}\). The randomly initialized model's errors are 11.2--13.5 times as large. Even for 100-dimensional instances beyond the training range, the trained UIR's mean validation MSE rises only to \(1.2\times10^{-2}\). This is approximately one-sixth of that of the randomly initialized model. This comparison shows that training enables generated UIR instances to better retain the objective function characteristics of MMMP. Together, the visualizations and matching results support the similarity between generated and real instances in both objective-space distributions and objective functions.

The comparison with NoInsGen in Table~\ref{tab:pap_perf} evaluates the benefit of incorporating UIR-based instance generation into PAP construction. NoInsGen retains DACMO's LLM-based algorithm generation and automatic parameter configuration but constructs PAPs directly on the initial real training instances. On MCCP, DACMO increases the aggregate mean normalized HV from 1.734 to 1.840, a gain of 6.2\%. On MMMP, the corresponding increase is from 1.658 to 1.715, or 3.4\%. The gains on MKP and MCIMP are 1.0\% and 0.3\%, respectively. The aggregate improvements are statistically significant on all four classes. These results demonstrate that UIR can learn from five real instances per class to provide effective training augmentation, improving the performance of the constructed PAPs on unseen real instances. Appendix~C of the supplementary material provides visualizations of augmented UIR instances and real test instances for the three bi-objective classes, together with MMMP matching results for augmented UIR instances.

\subsection{Effects of Co-Evolution and Automatic Algorithm Design}
\label{subsec:coevolution_algorithm_design}

\begin{figure*}[!t]
    \centering
    \subfloat[]{\includegraphics[width=0.243\textwidth]{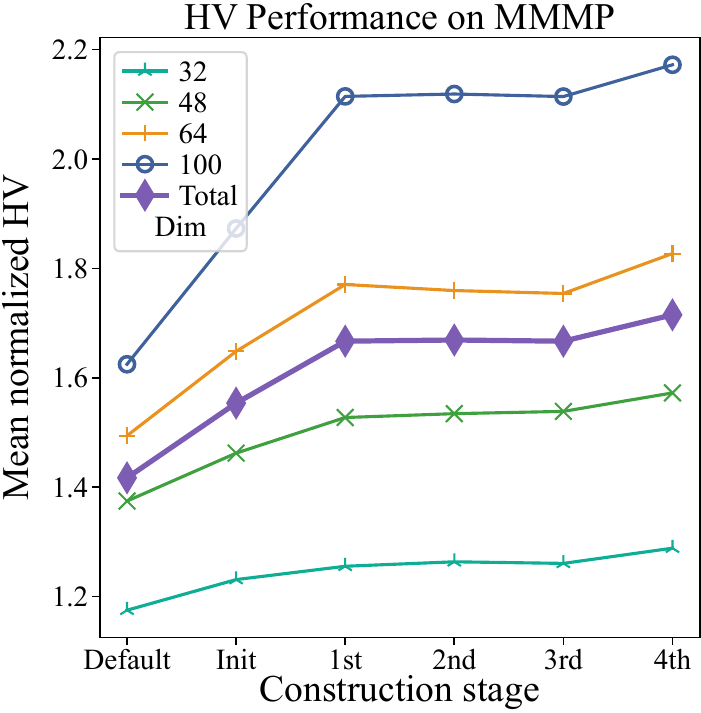}\label{fig:MMMP_iter}}
    \hfill
    \subfloat[]{\includegraphics[width=0.243\textwidth]{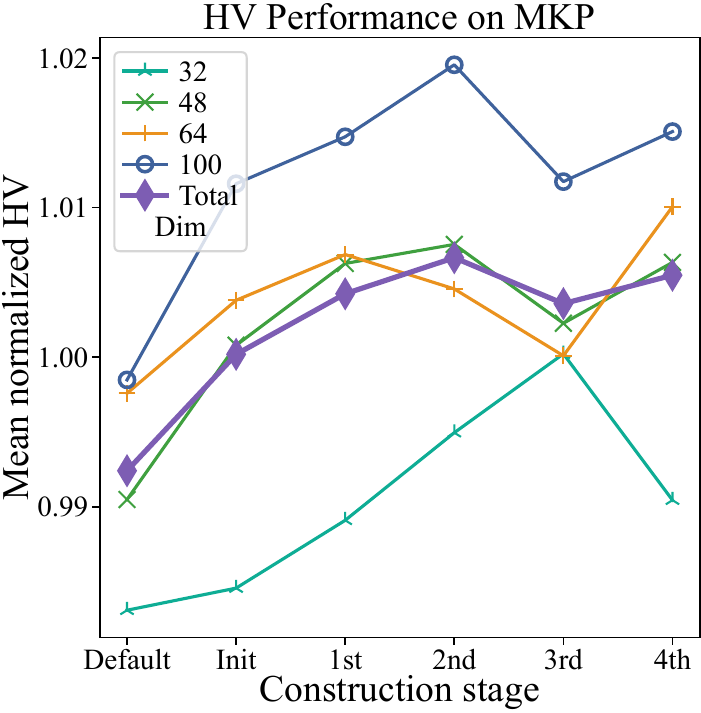}\label{fig:MKP_iter}}
    \hfill
    \subfloat[]{\includegraphics[width=0.243\textwidth]{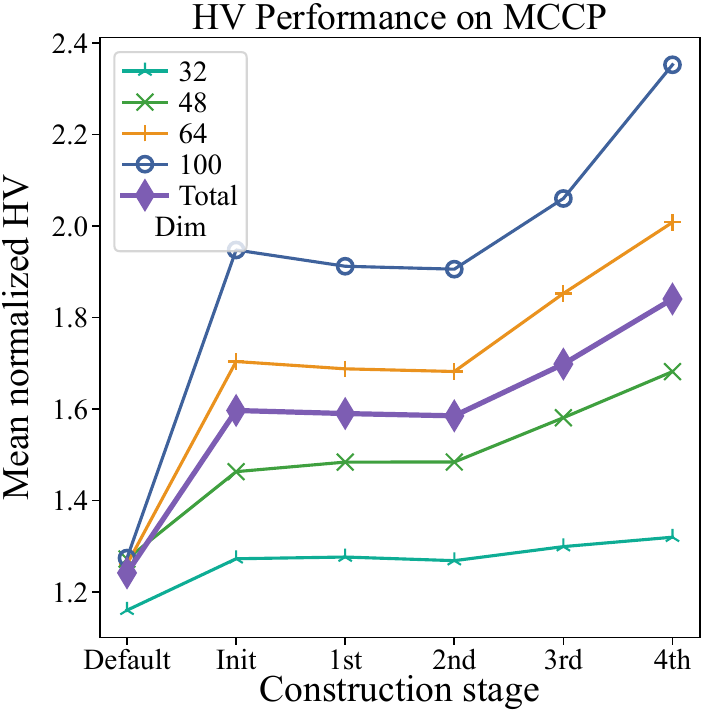}\label{fig:MCCP_iter}}
    \hfill
    \subfloat[]{\includegraphics[width=0.243\textwidth]{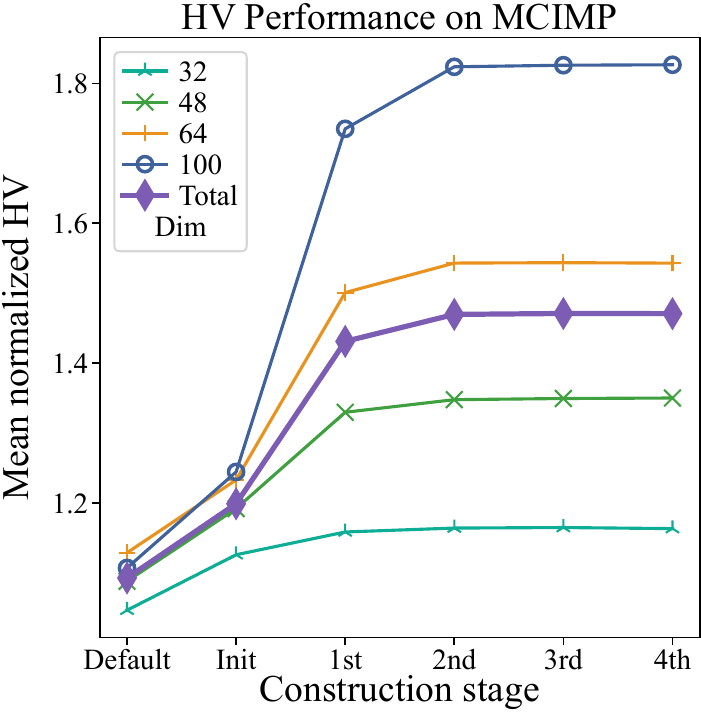}\label{fig:MCIMP_iter}}
    \caption{Mean normalized HV of PAPs on unseen real test instances at six construction stages: Default, Init, and four co-evolution rounds. The dimension-specific curves average over 30 instances each, and Total averages over all 120 instances of a problem class. (a) MMMP. (b) MKP. (c) MCCP. (d) MCIMP.}
    \label{fig:dacmo_iter}
\end{figure*}

To address \textit{RQ3}, we first evaluate the generalization gains from co-evolution relative to the initialized PAPs. Fig.~\ref{fig:dacmo_iter} reports test performance at six construction stages: Default, Init, and the four co-evolution rounds. Default comprises the default configurations of the four classical MOEAs, whereas Init denotes the PAP obtained after initial parameter tuning. All PAPs are evaluated on the same real test instances with the same FE budget per member. Each subplot corresponds to one problem class and shows the mean normalized HV at each test dimension, together with the aggregate mean across dimensions, labeled Total.

The final PAPs achieve higher aggregate test performance than Init on all four classes. Continued co-evolution yields substantial additional gains on MCCP in the third and fourth rounds and further improves PAP performance on MMMP after the early gains. On MCIMP, PAP performance improves mainly in the early rounds and remains stable thereafter. These results show that DACMO improves PAP generalization by alternating PAP updates with UIR-based instance augmentation.

Within this co-evolutionary process, automatic algorithm design expands the search strategies available for PAP updates. Its contribution is evaluated against NoAlgDesign in Table~\ref{tab:pap_perf}. NoAlgDesign uses DACMO's full augmented training set from the final round and constructs a PAP by tuning the four classical MOEAs with SMAC3. The comparison thus examines the additional value of algorithm design when UIR-generated training data are already available.

Automatic algorithm design delivers its largest gains on the two black-box problem classes from real-world applications. Relative to NoAlgDesign, DACMO increases aggregate mean normalized HV by 15.4\% on MCCP and 17.3\% on MCIMP. It also achieves a 3.8\% gain on the three-objective MMMP. The differences are statistically significant on all three classes. On MKP, the two PAPs have similar aggregate performance, showing that DACMO also remains competitive on this established benchmark. Automatic algorithm design thus strengthens PAP updates within the co-evolutionary process supported by UIR, enabling further generalization gains beyond parameter tuning of fixed algorithm templates.

\subsection{PAPs as Candidate Pools for Automatic Algorithm Selection}
\label{subsec:aas_candidate_pools}

\begin{table*}[tbp]
    \centering
    \caption{Normalized HV (Mean $\pm$ Std.) for DACMO PAPs under three AAS scenarios and standalone MOEAs. Oracle, Worst, and Random denote the per-instance maximum, minimum, and mean member performance, respectively. \textsuperscript{*}Oracle provides an upper bound and is excluded from best-result highlighting and statistical comparisons. The highest compared mean before rounding is \textbf{bolded} with a \colorbox{gray!70}{gray} background. \underline{Underlining} indicates no significant difference from that result. A superscript \(\dagger\) indicates that the marked Worst result is significantly better than the results of all six standalone MOEAs in the same row. Statistical comparisons use two-sided Wilcoxon tests with Holm correction at $\alpha=0.05$.}
    \label{tab:aas_perf}
    {
    \setlength{\aboverulesep}{0.05pt}
    \setlength{\belowrulesep}{0.05pt}
    \renewcommand{\arraystretch}{1.1}
    \setlength{\tabcolsep}{2.8pt}
    \begin{tabular*}{\textwidth}{@{\extracolsep{\fill}}ccccccccccc@{}}
        \toprule
        \multirow{2}[1]{*}{Problem} & \multirow{2}[1]{*}{Dim} & \multicolumn{3}{c}{DACMO} & \multirow{2}[1]{*}{NSGA-II} & \multirow{2}[1]{*}{NSGA-III} & \multirow{2}[1]{*}{SPEA2} & \multirow{2}[1]{*}{MOEA/D} & \multirow{2}[1]{*}{AGE-MOEA-II} & \multirow{2}[1]{*}{DCEA} \\
        \cmidrule{3-5}
         & & Oracle\textsuperscript{*} & Worst & Random & & & & & & \\
        \midrule
        \multirow{5}[1]{*}{MMMP} & 32 & 1.289±0.047 & 1.241±0.050\textsuperscript{\(\dagger\)} & \cellcolor{gray!70}\textbf{1.266±0.049} & 1.118±0.048 & 1.172±0.058 & 1.112±0.045 & 1.082±0.041 & 1.172±0.056 & 0.887±0.056 \\
        \cmidrule{2-11}
         & 48 & 1.572±0.109 & 1.486±0.097\textsuperscript{\(\dagger\)} & \cellcolor{gray!70}\textbf{1.532±0.104} & 1.184±0.066 & 1.375±0.098 & 1.155±0.060 & 1.127±0.065 & 1.383±0.092 & 0.879±0.087 \\
        \cmidrule{2-11}
         & 64 & 1.827±0.102 & 1.706±0.107\textsuperscript{\(\dagger\)} & \cellcolor{gray!70}\textbf{1.771±0.102} & 1.250±0.054 & 1.494±0.085 & 1.183±0.055 & 1.153±0.057 & 1.574±0.077 & 0.902±0.090 \\
        \cmidrule{2-11}
         & 100 & 2.172±0.178 & 1.979±0.150\textsuperscript{\(\dagger\)} & \cellcolor{gray!70}\textbf{2.074±0.158} & 1.263±0.066 & 1.625±0.109 & 1.222±0.068 & 1.160±0.047 & 1.807±0.127 & 0.891±0.116 \\
        \cmidrule{2-11}
         & Total & 1.715±0.346 & 1.603±0.293\textsuperscript{\(\dagger\)} & \cellcolor{gray!70}\textbf{1.661±0.318} & 1.204±0.082 & 1.416±0.189 & 1.168±0.070 & 1.131±0.061 & 1.484±0.252 & 0.890±0.090 \\
        \midrule
        \multirow{5}[1]{*}{MKP} & 32 & 0.990±0.025 & 0.945±0.025\phantom{\textsuperscript{\(\dagger\)}} & \underline{0.966±0.023} & \underline{0.959±0.031} & \cellcolor{gray!70}\textbf{0.971±0.020} & \underline{0.961±0.023} & \underline{0.970±0.026} & \underline{0.964±0.040} & 0.917±0.036 \\
        \cmidrule{2-11}
         & 48 & 1.006±0.032 & 0.950±0.043\phantom{\textsuperscript{\(\dagger\)}} & \cellcolor{gray!70}\textbf{0.977±0.027} & \underline{0.966±0.023} & \underline{0.973±0.032} & \underline{0.963±0.037} & \underline{0.963±0.035} & \underline{0.977±0.027} & 0.921±0.028 \\
        \cmidrule{2-11}
         & 64 & 1.010±0.029 & 0.962±0.023\phantom{\textsuperscript{\(\dagger\)}} & \cellcolor{gray!70}\textbf{0.986±0.022} & \underline{0.979±0.026} & \underline{0.981±0.022} & \underline{0.974±0.019} & \underline{0.982±0.023} & \underline{0.982±0.024} & 0.933±0.028 \\
        \cmidrule{2-11}
         & 100 & 1.015±0.025 & 0.975±0.021\phantom{\textsuperscript{\(\dagger\)}} & \underline{0.994±0.022} & 0.982±0.015 & \underline{0.985±0.022} & 0.980±0.019 & \underline{0.985±0.024} & \cellcolor{gray!70}\textbf{0.996±0.023} & 0.942±0.027 \\
        \cmidrule{2-11}
         & Total & 1.006±0.029 & 0.958±0.032\phantom{\textsuperscript{\(\dagger\)}} & \cellcolor{gray!70}\textbf{0.981±0.026} & 0.972±0.026 & \underline{0.978±0.025} & 0.970±0.027 & \underline{0.975±0.029} & \underline{0.980±0.032} & 0.928±0.032 \\
        \midrule
        \multirow{5}[1]{*}{MCCP} & 32 & 1.320±0.042 & 1.290±0.043\textsuperscript{\(\dagger\)} & \cellcolor{gray!70}\textbf{1.305±0.042} & 1.058±0.046 & 1.146±0.051 & 1.129±0.042 & 1.083±0.037 & 1.147±0.048 & 0.935±0.062 \\
        \cmidrule{2-11}
         & 48 & 1.682±0.095 & 1.547±0.103\textsuperscript{\(\dagger\)} & \cellcolor{gray!70}\textbf{1.624±0.087} & 1.148±0.092 & 1.238±0.088 & 1.201±0.076 & 1.159±0.045 & 1.269±0.082 & 1.103±0.073 \\
        \cmidrule{2-11}
         & 64 & 2.008±0.147 & 1.773±0.141\textsuperscript{\(\dagger\)} & \cellcolor{gray!70}\textbf{1.898±0.123} & 1.098±0.093 & 1.242±0.094 & 1.163±0.089 & 1.137±0.081 & 1.265±0.092 & 1.280±0.095 \\
        \cmidrule{2-11}
         & 100 & 2.352±0.210 & 1.967±0.171\textsuperscript{\(\dagger\)} & \cellcolor{gray!70}\textbf{2.163±0.164} & 1.087±0.083 & 1.201±0.117 & 1.207±0.098 & 1.115±0.081 & 1.285±0.123 & 1.585±0.114 \\
        \cmidrule{2-11}
         & Total & 1.840±0.407 & 1.644±0.282\textsuperscript{\(\dagger\)} & \cellcolor{gray!70}\textbf{1.747±0.338} & 1.097±0.087 & 1.207±0.099 & 1.175±0.085 & 1.124±0.070 & 1.241±0.106 & 1.226±0.256 \\
        \midrule
        \multirow{5}[1]{*}{MCIMP} & 32 & 1.164±0.126 & 1.149±0.119\textsuperscript{\(\dagger\)} & \cellcolor{gray!70}\textbf{1.157±0.122} & 1.010±0.086 & 1.030±0.124 & 1.006±0.109 & 0.929±0.106 & 1.046±0.130 & 1.060±0.099 \\
        \cmidrule{2-11}
         & 48 & 1.350±0.205 & 1.334±0.202\textsuperscript{\(\dagger\)} & \cellcolor{gray!70}\textbf{1.342±0.203} & 1.023±0.085 & 1.082±0.117 & 1.017±0.076 & 0.977±0.057 & 1.099±0.102 & 1.201±0.172 \\
        \cmidrule{2-11}
         & 64 & 1.543±0.291 & 1.519±0.283\textsuperscript{\(\dagger\)} & \cellcolor{gray!70}\textbf{1.531±0.287} & 1.049±0.062 & 1.120±0.098 & 1.045±0.068 & 0.998±0.049 & 1.139±0.136 & 1.353±0.222 \\
        \cmidrule{2-11}
         & 100 & 1.827±0.480 & 1.795±0.464\textsuperscript{\(\dagger\)} & \cellcolor{gray!70}\textbf{1.812±0.472} & 1.022±0.066 & 1.097±0.137 & 1.017±0.058 & 0.962±0.072 & 1.133±0.163 & 1.574±0.348 \\
        \cmidrule{2-11}
         & Total & 1.471±0.392 & 1.449±0.380\textsuperscript{\(\dagger\)} & \cellcolor{gray!70}\textbf{1.461±0.386} & 1.026±0.077 & 1.082±0.124 & 1.021±0.081 & 0.967±0.078 & 1.104±0.140 & 1.297±0.298 \\
        \bottomrule
    \end{tabular*}
    }
\end{table*}

For each target MOBOP class, DACMO constructs a PAP from a small set of real training instances for direct use on unseen instances. Its members can also serve as candidate configurations for AAS, enabling per-instance selection without further algorithm design or parameter tuning for each new instance. To address \textit{RQ4}, Table~\ref{tab:aas_perf} compares the normalized HV under Oracle, Worst, and Random selection with that of six standalone MOEAs. Each PAP member and standalone algorithm is evaluated with the same budget of \(\#\mathrm{FEs}=1600\).

The constructed pools deliver strong performance even under worst-case selection on unseen instances. On MMMP, MCCP, and MCIMP, Worst significantly outperforms all six standalone MOEAs at every tested dimension and in the aggregate results. Its aggregate mean normalized HV reaches 1.603, 1.644, and 1.449, respectively. These values exceed the highest aggregate mean achieved by a standalone MOEA in each class by 8.0\%, 32.4\%, and 11.8\%, respectively. The overall advantage on these classes therefore does not depend on accurately selecting the best member for each unseen instance.

Random further demonstrates the value of directly reusing the constructed configurations on unseen instances. Relative to the best standalone MOEA in each class, Random improves aggregate mean normalized HV by 11.9\% on MMMP, 40.8\% on MCCP, and 12.6\% on MCIMP. Its performance is significantly better than that of all six standalone MOEAs on these three classes. On MKP, Random achieves an aggregate mean normalized HV of 0.981, also exceeding the means of all six standalone MOEAs. These expected gains can therefore be obtained with a single member's evaluation budget, without using instance information to guide selection.

Oracle gives the ideal performance attainable by selecting the best member for each instance. The gap between Oracle and Random indicates the potential benefit of improved member selection. Together, the Worst and Random results demonstrate that DACMO can construct high-quality AAS candidate pools from a small set of real training instances. Their member configurations can be directly reused on unseen instances of the target MOBOP class.

\section{Conclusion}
\label{sec:conclusion}

This work proposes DACMO, the first general-purpose method for automatic PAP construction on MOBOPs. Its core, UIR, represents and generates multi-objective binary instances with varying dimensions without knowledge of problem-specific instance structures. During generation, UIR preserves common features learned from real instances of the target class. DACMO combines UIR-based instance augmentation with automatic algorithm design through co-evolution. This enables the construction of generalizable PAPs without expert customization of instance generators or search algorithm templates for the target class.

Experiments on four MOBOP classes demonstrate that DACMO constructs effective PAPs from five real training instances per class. Overall, these PAPs outperform portfolios of tuned classical MOEAs and those constructed by EoH-S. UIR-based instance augmentation significantly improves PAP generalization on all four classes. Without problem-specific generator customization, DACMO also achieves competitive performance against CEPS, which relies on expert-designed generators, and significantly outperforms it on two classes. The resulting PAPs can be directly applied to unseen instances of the target class, including those beyond the training dimension range, and provide high-quality candidate pools for AAS.

\bibliographystyle{IEEEtran}
\bibliography{DACMO}

\clearpage
\twocolumn[{
    \centering
    {\Huge Supplementary for ``Constructing Generalizable Parallel Algorithm Portfolios for Multi-Objective Binary Optimization via General-Purpose Instance Generation''}\\[3em]
}]
\setcounter{algocf}{2}
\setcounter{figure}{4}
\setcounter{table}{5}
\appendices
\section{UIR-Based Instance Generation Using PGPE}
\label{app:pgpe}

PGPE~\cite{sehnke_parameter-exploring_2010} searches instance embeddings to generate UIR instances with low normalized performance for the current PAP. It implements the search objective in~(8) of the main paper and is used during training set evolution in Section~IV-C. Given a starting UIR instance \(\tilde{s}_0\), the search initializes its mean at the instance embedding \(\mathbf{e}_0\). The learned shared parameters \(\mathbf{w}_E\), \(\mathbf{w}_D\), and \(\mathbf{w}_H\) remain fixed. All instances generated in this search have the same number of decision variables as \(\tilde{s}_0\).

Algorithm~\ref{alg:pgpe} gives the sampling and update procedure. Each iteration evaluates the center of the search distribution and \(N\) pairs of symmetrically sampled embeddings (lines~\ref{line:pgpe_sample}--\ref{line:pgpe_evaluate}). Each embedding defines a UIR instance, whose difficulty is measured by \(\bar f(P,\tilde{s})\) from~(7) of the main paper. PGPE uses rewards obtained from the negative normalized performance, so that larger rewards favor instances on which \(P\) performs worse. The rewards are converted to centered ranks before updating the search distribution. Specifically, the ranks of the \(2N+1\) rewards range from 0 to \(2N\), and each rank is divided by \(2N\) and shifted by \(-1/2\). The search retains the instance with the lowest normalized performance observed across iterations.

The perturbation matrices \(\mathbf{E},\mathbf{S}\in\mathbb{R}^{N\times d_e}\) in Algorithm~\ref{alg:pgpe} are used to update the mean and standard deviation vectors. The reward difference vector \(\mathbf{a}\in\mathbb{R}^{N}\) guides the mean update, while \(\mathbf{b}\in\mathbb{R}^{N}\) compares the average reward of each symmetric pair with the center reward. The scalar \(s_r\) is the population standard deviation of the \(2N\) sampled rewards, excluding the center. The operation \(\operatorname{clip}\) applies its bounds componentwise. Standard deviation updates are bounded relative to their current values, followed by gradual decay of the exploration scale and the mean learning rate.

In all four problem classes, \(N=40\), giving 81 evaluated embeddings per iteration, and the maximum number of iterations is \(L=200\). The initial standard deviation vector is \(\boldsymbol{\upsigma}_{\mathrm{init}}=\mathbf{1}\), the initial mean learning rate is \(\alpha_{\mu}=0.05\), and the standard deviation learning rate is \(\alpha_{\sigma}=0.1\). The decay factors are \(\gamma_{\sigma}=0.999\) and \(\gamma_{\mu}=0.9999\). Decay is applied when the corresponding standard deviation exceeds \(\sigma_{\mathrm{lim}}=0.01\) or the mean learning rate exceeds \(\alpha_{\mu,\mathrm{lim}}=0.001\).

The search also uses the cached performance values \(\mathcal{G}=\{\bar f(P,\tilde{s}):\tilde{s}\in M_{\mathrm{pool}}\}\) from the current comparison pool. With iterations indexed from zero, it can stop when the best value found is no greater than \(\min\mathcal{G}\) after \(\lfloor L/8\rfloor\) iterations, or no greater than the mean of \(\mathcal{G}\) after \(\lfloor L/4\rfloor\) iterations (line~\ref{line:pgpe_stop}). The resulting instance is then passed to the augmentation step in Section~IV-C of the main paper.

\begin{algorithm}[htbp]
    \small
    \caption{PGPE Search for a UIR Instance}
    \label{alg:pgpe}
    \SetKwInput{KwData}{Input}
    \SetKwInput{KwResult}{Output}
    \SetKw{To}{to}
    \SetKw{Break}{break}
    \KwData{Starting UIR instance \(\tilde{s}_0\) with embedding \(\mathbf{e}_0\), PAP \(P\), cached comparison-pool performance values \(\mathcal{G}\), number \(N\) of symmetric sample pairs, iteration limit \(L\), and PGPE hyperparameters.}
    \KwResult{Generated UIR instance \(\tilde{s}'\).}
    \(\boldsymbol{\upmu}\leftarrow\mathbf{e}_0,\quad \boldsymbol{\upsigma}\leftarrow\boldsymbol{\upsigma}_{\mathrm{init}},\quad \bar f_{\mathrm{best}}\leftarrow+\infty\)\;
    Keep \(P\), the shared UIR parameters, and the decision-variable dimension of \(\tilde{s}_0\) fixed\;
    \For{\(\ell\leftarrow 0\) \To \(L-1\)}{
        Sample \(\boldsymbol{\upepsilon}_i\sim\mathcal{N}\!\left(\mathbf{0},\operatorname{diag}(\boldsymbol{\upsigma}^{2})\right)\), \(i=1,\ldots,N\)\nllabel{line:pgpe_sample}\;
        \(\mathbf{e}_0\leftarrow\boldsymbol{\upmu},\quad \mathbf{e}_i\leftarrow\boldsymbol{\upmu}+\boldsymbol{\upepsilon}_i,\quad \mathbf{e}_{N+i}\leftarrow\boldsymbol{\upmu}-\boldsymbol{\upepsilon}_i\), \(i=1,\ldots,N\)\;
        Form \(\tilde{s}_i=\tilde{s}(\mathbf{e}_i)\) and evaluate \(\bar f_i\leftarrow\bar f(P,\tilde{s}_i)\), \(i=0,\ldots,2N\)\nllabel{line:pgpe_evaluate}\;
        \(k\leftarrow\operatorname*{arg\,min}_{0\leq i\leq 2N}\bar f_i\)\;
        \lIf{\(\bar f_k<\bar f_{\mathrm{best}}\)}{\(\tilde{s}'\leftarrow\tilde{s}_k,\quad\bar f_{\mathrm{best}}\leftarrow\bar f_k\)}
        \((r_0,\ldots,r_{2N})\leftarrow\operatorname{CenteredRanks}(-\bar f_0,\ldots,-\bar f_{2N})\)\;
        \(E_{ij}\leftarrow\epsilon_{ij},\quad S_{ij}\leftarrow(\epsilon_{ij}^{2}-\sigma_j^{2})/\sigma_j\), \(i=1,\ldots,N\), \(j=1,\ldots,d_e\)\;
        \(a_i\leftarrow r_i-r_{N+i},\quad b_i\leftarrow(r_i+r_{N+i})/2-r_0\), \(i=1,\ldots,N\)\;
        \(s_r\leftarrow\operatorname{Std}(r_1,\ldots,r_{2N})\)\;
        \(\boldsymbol{\upmu}\leftarrow\boldsymbol{\upmu}+\alpha_{\mu}\mathbf{E}^{\mathsf T}\mathbf{a}\)\;
        \(\mathbf{u}\leftarrow\dfrac{\alpha_{\sigma}}{2Ns_r}\mathbf{S}^{\mathsf T}\mathbf{b}\)\;
        \(\boldsymbol{\upsigma}\leftarrow\boldsymbol{\upsigma}+\operatorname{clip}(\mathbf{u},-\boldsymbol{\upsigma}/2,\boldsymbol{\upsigma})\)\;
        \(\sigma_j\leftarrow\gamma_{\sigma}\sigma_j\) for each \(j\) with \(\sigma_j>\sigma_{\mathrm{lim}}\)\;
        \lIf{\(\alpha_{\mu}>\alpha_{\mu,\mathrm{lim}}\)}{\(\alpha_{\mu}\leftarrow\gamma_{\mu}\alpha_{\mu}\)}
        \lIf{\(\bigl(\ell\geq\lfloor L/8\rfloor\ \text{and}\ \bar f_{\mathrm{best}}\leq\min\mathcal{G}\bigr)\) or \(\bigl(\ell\geq\lfloor L/4\rfloor\ \text{and}\ \bar f_{\mathrm{best}}\leq\operatorname{Mean}(\mathcal{G})\bigr)\)}{\Break\nllabel{line:pgpe_stop}}
    }
    \Return{\(\tilde{s}'\)}\;
\end{algorithm}

\clearpage
\section{Prompts and Code Template for LLM-Based Algorithm Generation}
\label{app:prompt}

\subsection{Objective Direction and MOEA Interface}

The main paper formulates the MOBOPs as maximization problems, whereas the MOEA implementation uses a minimization interface. Objective vectors are therefore multiplied by \(-1\) before they are passed to the solvers. Minimizing these transformed objectives is equivalent to maximizing the original objectives and preserves the corresponding Pareto-optimal solution set. The minimization instructions in the prompts below follow this implementation convention.

The MOEA template in Algorithm~2 of the main paper fixes the evaluation interface and execution loop. The LLM supplies the initialization, offspring generation, and population update operators, together with their tunable parameters and ranges. The following excerpt gives the corresponding implementation interfaces. The method \texttt{gen\_offsprings} implements the offspring generation step in Algorithm~2. The inherited \texttt{evaluate} method counts function evaluations and records the evaluated solutions. The inherited \texttt{get\_known\_solutions} method returns these solutions and their objective vectors, from which the nondominated output set is extracted.

\begin{rawbox}
class MOEA:
    def initialization(self) -> Tuple[List[NpArray], NpArray]:
        pass

    def gen_offsprings(
        self, curr_pop: List[NpArray], curr_objs: NpArray
    ) -> Tuple[List[NpArray], NpArray]:
        pass

    def update(
        self, curr_pop: List[NpArray], offsprings: List[NpArray],
        curr_objs: NpArray, offspring_objs: NpArray
    ) -> Tuple[List[NpArray], NpArray]:
        pass

    def run(self):
        curr_pop, curr_objs = self.initialization()
        while self.eval_time < self.max_eval:
            offsprings, offspring_objs = self.gen_offsprings(
                curr_pop, curr_objs
            )
            curr_pop, curr_objs = self.update(
                curr_pop, offsprings, curr_objs, offspring_objs
            )
        return self.get_known_solutions()
\end{rawbox}

\subsection{LLM Prompt Templates}

As described in Section~IV-B of the main paper, Kimi-K2.5 receives the current PAP members' code, parameter settings, and performance feedback to generate a new algorithm and its configuration space. The response contains the algorithm code, its description, and the parameter definitions in a single JSON object. The system prompt specifies this output structure:

\begin{rawbox}
The output result must contain only one JSON, and the JSON template is:
```json
{
  "hyperparameters": [
    {
      "name": "hyperparameter name",
      "type": "hyperparameter type. The value of this field should be a string. The value can only be: \"int\" or \"float\" or \"bool\"",
      "default": "default value of hyperparameter. The type of the value in this field needs to correspond to the type of the hyperparameter, not be of the str type.",
      "categorical": "true or false, the hyperparameter is categorical or not. The value of this field should be a bool value. It should be noted that even for variables of type int, if their optional values are within a range rather than just a few specific values, then categorical should be set to false. For example, population size is an integer, but since its value can be any integer within a range, categorical should be false.",
      "range": "range of the hyperparameter, list with length 2 if hyperparameter is not categorical, set if hyperparameter is categorical",
      "description": "description of the hyperparameter"
    }
  ],
  "code": " The program code of the algorithm ",
  "description": "description of algorithm"
}
```
\end{rawbox}

The user prompt combines the MOEA code template with a report of the current PAP members' performance. The values labeled \texttt{HV} in the report are normalized HV values as defined in~(7) of the main paper. The report also includes member rankings on individual instances and summaries by dimension. Ellipses indicate omitted code and repeated feedback entries. The complete output schema is given in the system prompt above.

\begin{rawbox}
There is a python file "moea.py" with the content following:
```Python
# moea.py
from src.types_ import *

class MOEA:
    def __init__(self, max_eval: int, objective_num: int, dim: int, bound_val: List[List[float]],
                 eval_func: Callable[[np.ndarray], np.ndarray], hyper_param: Dict):
        self.max_eval: int = max_eval
    
    ... 
    
    def run(self):
        curr_pop, curr_objs = self.initialization()
        while self.eval_time < self.max_eval:
            offsprings, offspring_objs = self.gen_offsprings(curr_pop, curr_objs)
            curr_pop, curr_objs = self.update(curr_pop, offsprings, curr_objs, offspring_objs)
        return self.get_known_solutions()
```

There is also a JSON file "json_template.json" with the content following:
```json
{
  "hyperparameters": [
  
  ...
  
  "description": "description of algorithm"
}
```
I need you to use the MOEA class in the given moea.py file as the parent class to implement a complete code for an optimization algorithm for 0-1 multi-objective optimization problems. The dimension of the optimization problem is 32-100, the number of objectives is 2-3, and all are minimization objectives. The algorithm needs to implement a class named NewMOEA that inherits from the MOEA class and implements the three methods initialization(), gen_offsprings(), and update(). The parameter lists of these methods must not be changed when implementing. At the same time, all hyperparameter lists required for implementation should be placed separately in MOEA.hyper_param. Do not modify the "__init__" function, "evaluate" function, and "run" function, but directly inherit the methods from the MOEA class! The statement to import the MOEA class is "from src.moea.moea import MOEA". The code also needs "from src.types_ import *" to import some custom type names.

When implementing the new optimization algorithm, you can analyze and design based on some currently implemented algorithms and their performance. The code, description, and performance of these algorithms using the listed hyperparameter values on a problem instance set used for verifying performance are as follows:

Performance Summary of Solvers

--- Solver 0 ---
Description: NSGA-II (Non-dominated Sorting Genetic Algorithm II) is a popular multi-objective evolutionary algorithm, ...
Code:
```Python
from src.moea.moea import MOEA
from src.types_ import *
import numpy as np
from typing import List, Tuple, Dict, Set
import random
class NewMOEA(MOEA):
    def _non_dominated_sort(self, objs: NpArray) -> Tuple[List[List[int]], NpArray]:
        """Fast non-dominated sorting algorithm."""
        n = len(objs)
        S: Dict[int, Set[int]] = {i: set() for i in range(n)}
...
    
```

Hyper-parameters:
```JSON
{"pop_size": 143, "crossover_rate": 0.8996722666924, "mutation_rate": 0.0198469733669, "tournament_size": 10}
```

HV Performance across all UIR instances:
  UIR instance 0 (dim=32): HV=0.890960, Rank=3
  UIR instance 1 (dim=40): HV=0.988416, Rank=1
...

Overall Rank Statistics:
  Rank 1: 2 times
  Rank 2: 1 times
...

--- Solver 1 ---

...

--- Solver 2 ---

...

--- Solver 3 ---

...

Rank Statistics by Dimension

--- Dimension 32 ---
  Solver 0:
    Rank 3: 1 times
...
--- Dimension 40 ---
...
--- Dimension 48 ---
...  
--- Dimension 56 ---
...
--- Dimension 64 ---
...

Please refer to existing methods and their performance to implement a complete code for an optimization algorithm for 0-1 multi-objective optimization problems. The dimension of the optimization problem is 32-100, the number of objectives is 2-3, and all are minimization objectives. The algorithm needs to implement a class named NewMOEA that inherits from the MOEA class. Please return the implementation result in JSON format according to the format in the json_template.json file. In the "hyperparameters" field, all required hyperparameters need to be listed separately with recommended default values and ranges. Hyperparameters can only be of int, float, or bool types. The "categorical" should indicate the hyperparameter is categorical or not. It should be noted that even for variables of type int, if their optional values are within a range rather than just a few specific values, then categorical should be set to false. For example, population size is an integer, but since its value can be any integer within a range, categorical should be false. For not categorical ranges (whether int or float), the "range" field should contain two elements representing the start and end of the hyperparameter range; for categorical ranges, the range field should contain all candidate values. The description of this algorithm should be placed in the "description" field. The complete code file of the implemented new algorithm should be placed in the "code" field.

Please note that the new algorithm must be implemented as described above. The parameter lists of the initialization(), gen_offsprings(), and update() methods must not be changed when implementing. Do not modify the "__init__" function, "evaluate" function, and run function, but directly inherit from the MOEA class. At the same time, all hyperparameter lists required for implementation should be placed separately in MOEA.hyper_param. All evaluations of the solution should be performed through the self.evaluate() method, and direct calls to self.eval_func are strictly prohibited. Do not modify the "__init__" function, "evaluate" function, and "run" function. These three functions must not appear in the generated code, otherwise it will be considered an error! Must return in the given JSON format! The complete code needs to be placed directly in the "code" field, not expressed in separate functions. The final JSON only contains one "code" field. Again, the implemented class name is NewMOEA, and the generated code must not contain the "__init__" function, "evaluate" function, and "run" function. Please ensure that the generated code is correct and meets the requirements of multi-objective minimization optimization algorithms. In addition, it is necessary to carefully check whether the generated code will cause array out-of-bounds errors. When determining whether a variable is None, the statement "if var is not None" must be used accurately but not "if var" only. All evaluations of the solution should be performed through the self.evaluate() method, and direct calls to self.eval_func are strictly prohibited.

The output MUST be ONLY a valid JSON string that strictly matches the template provided above. No extra text, explanation, or content is allowed.
\end{rawbox}

\clearpage
\section{Visualizations and Similarity Analysis of Augmented UIR Instances}
\label{app:test_obj_space}
\label{app:augment_mmmp_similarity}

\begin{figure*}[htbp]
    \centering
    \includegraphics[width=0.70\textwidth]{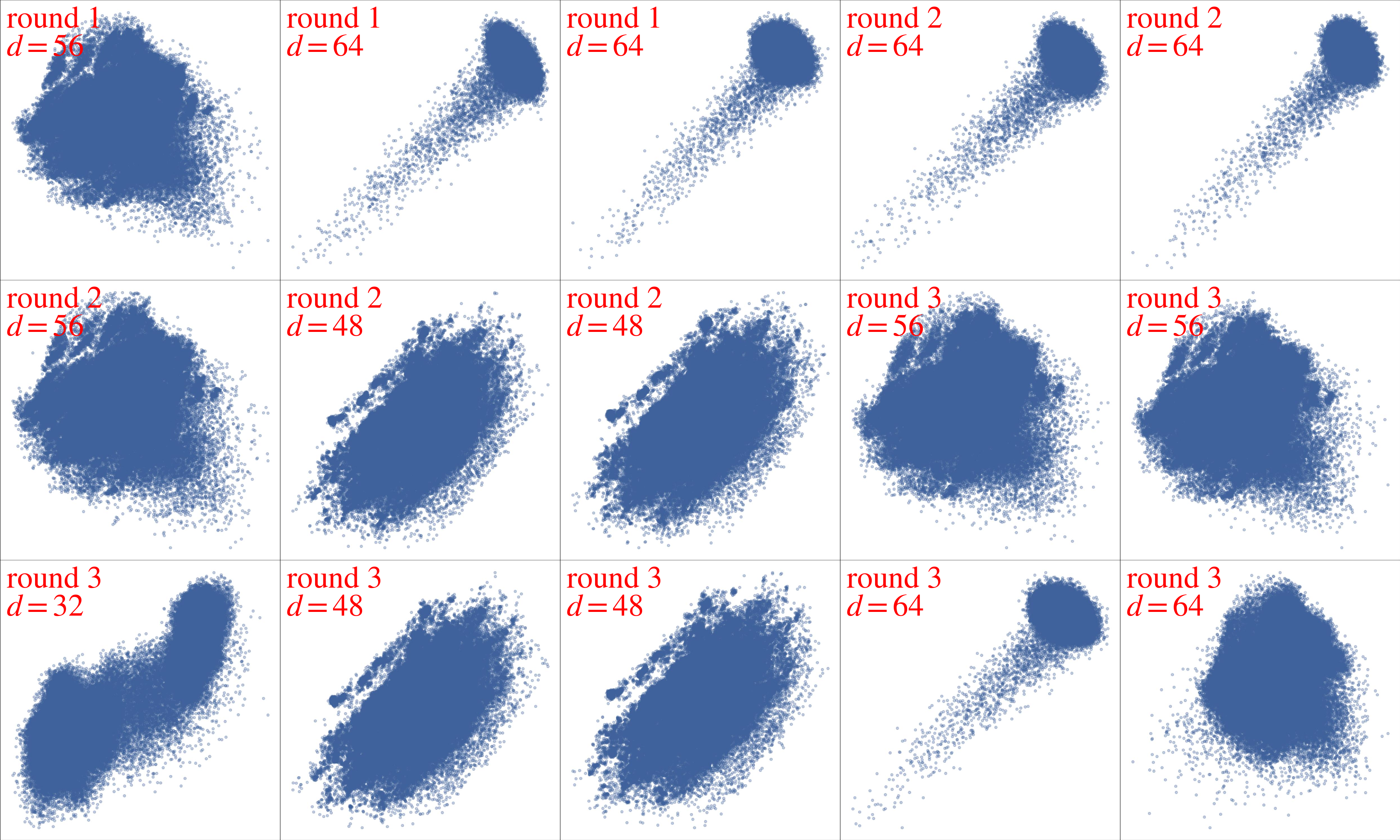}
    \caption{Objective-space distributions of the 15 UIR instances added to the MKP training set during DACMO's co-evolution. Each plot corresponds to one instance and is labeled with its augmentation round and decision-variable dimension \(d\).}
    \label{fig:MKP_augment}
    \par\medskip
    \subfloat[]{\includegraphics[width=0.40\textwidth]{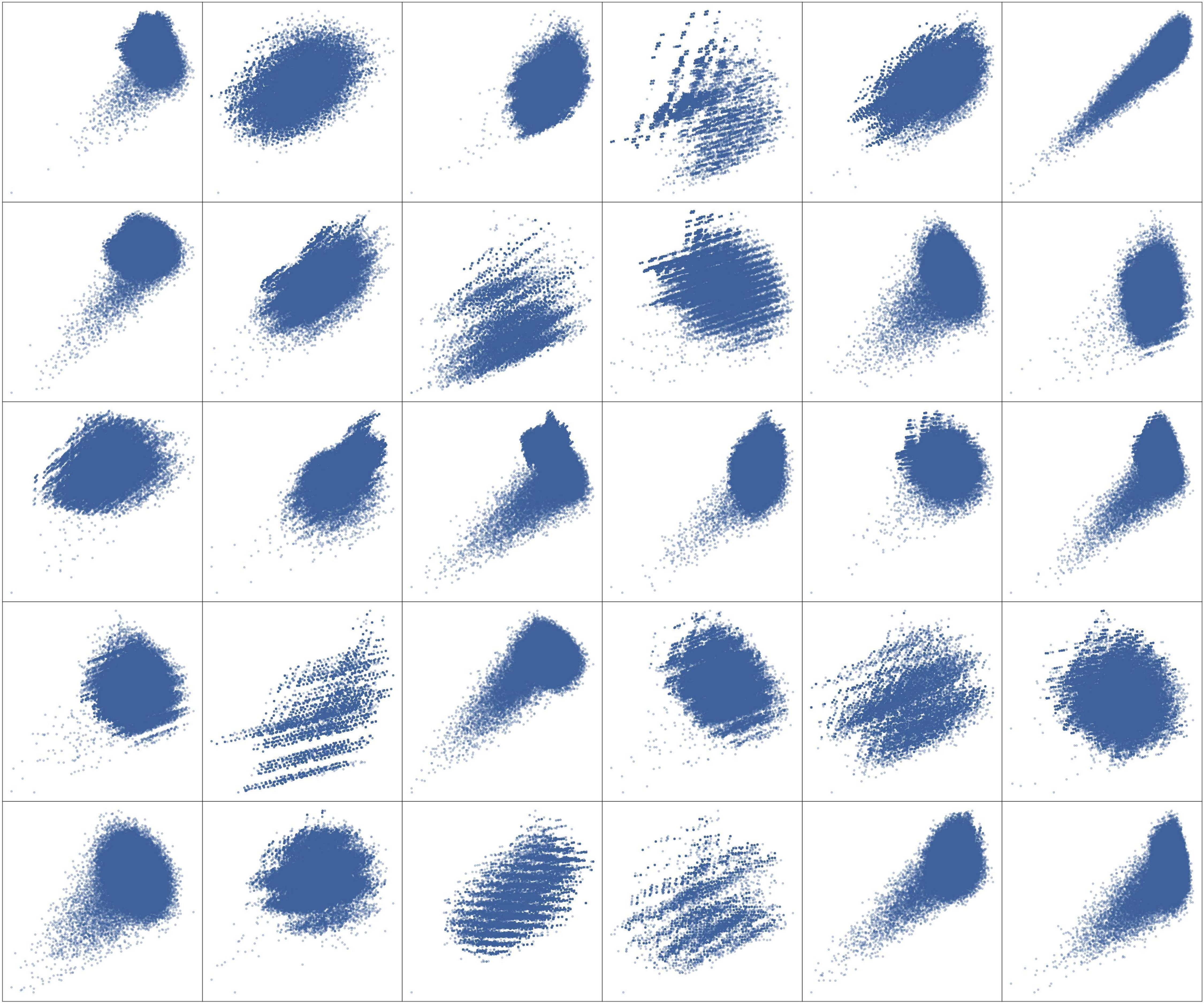}
    \label{fig:MKP_test_vis_32}}
    \subfloat[]{\includegraphics[width=0.40\textwidth]{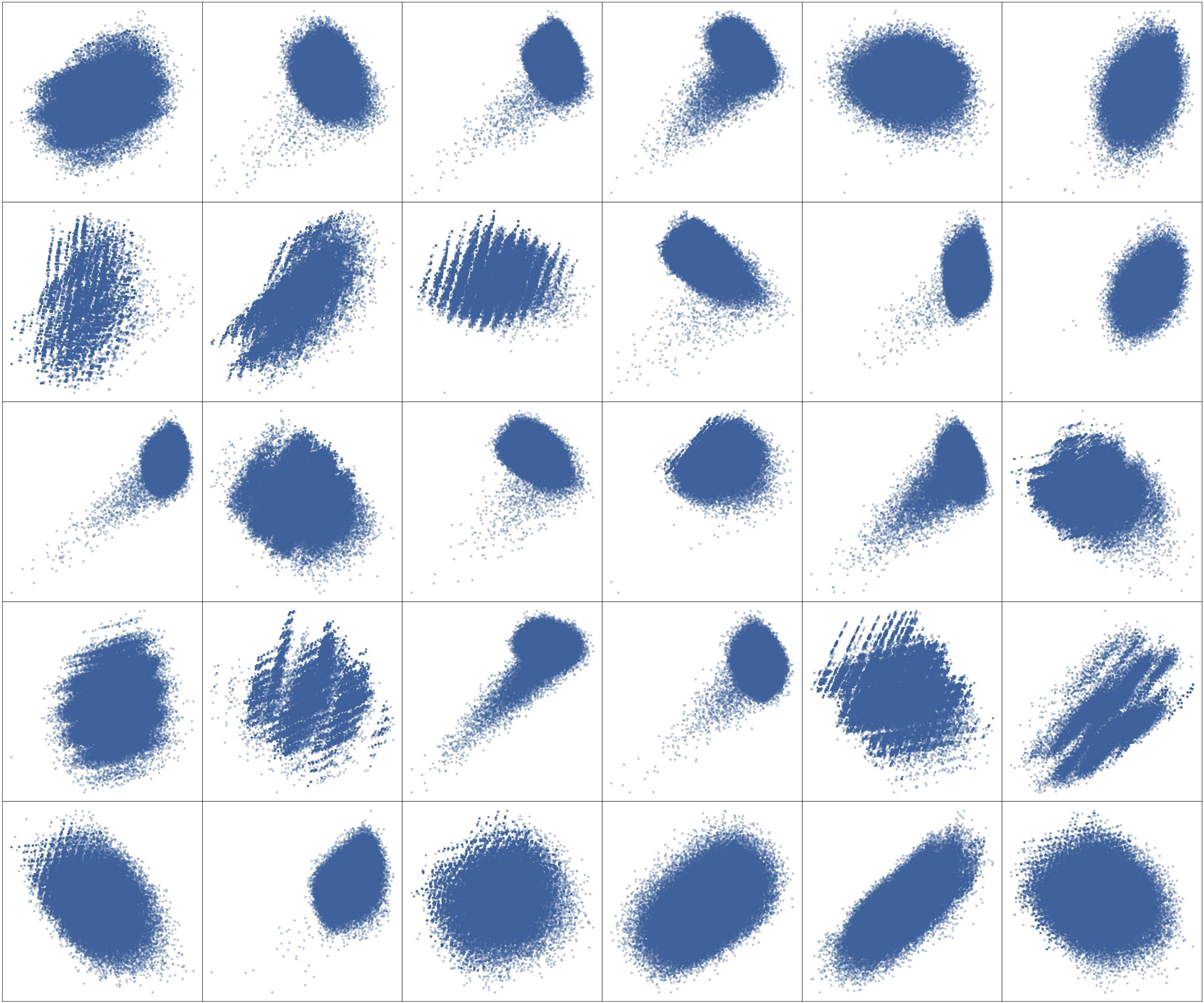}
    \label{fig:MKP_test_vis_48}}\\
    \subfloat[]{\includegraphics[width=0.40\textwidth]{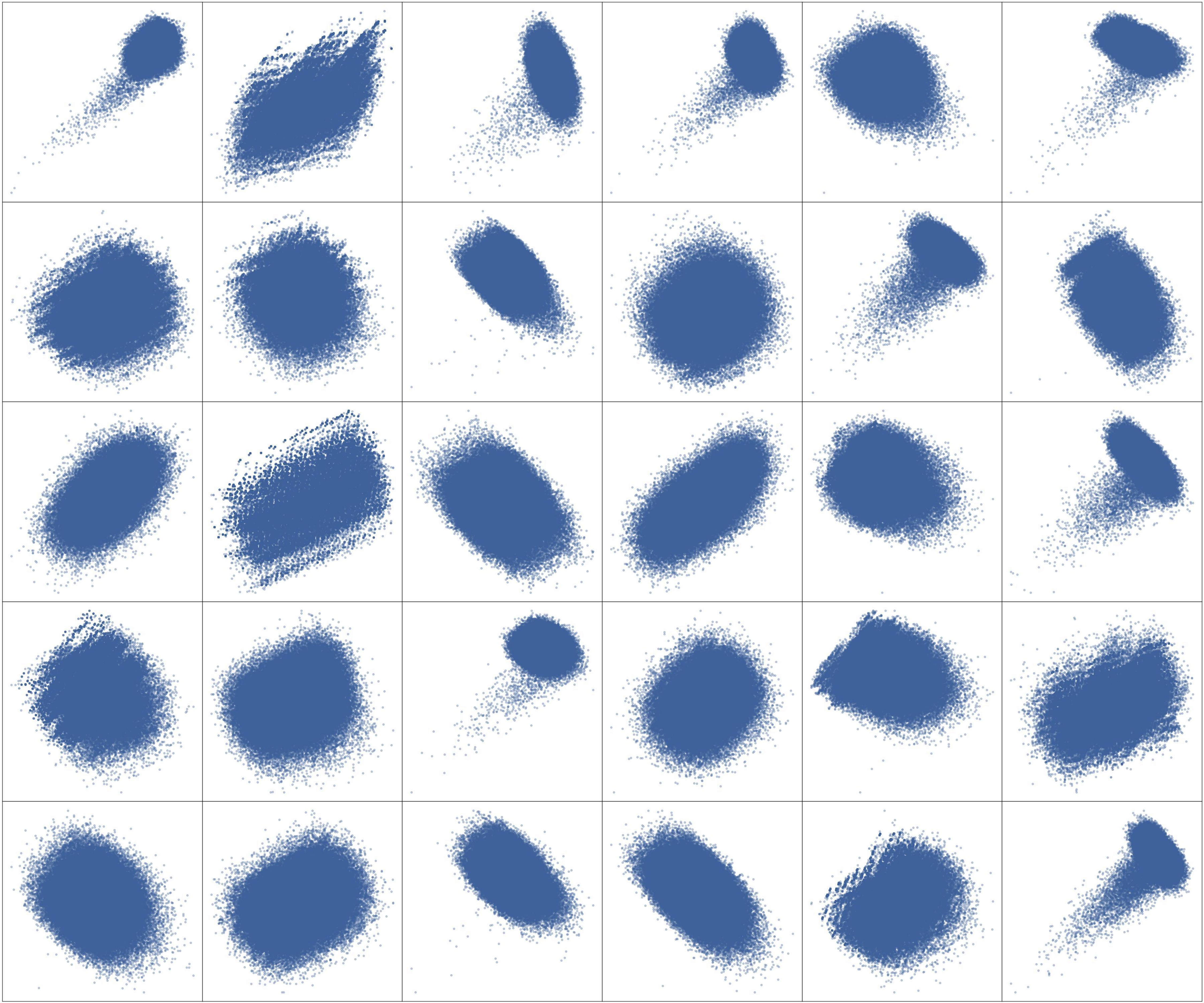}
    \label{fig:MKP_test_vis_64}}
    \subfloat[]{\includegraphics[width=0.40\textwidth]{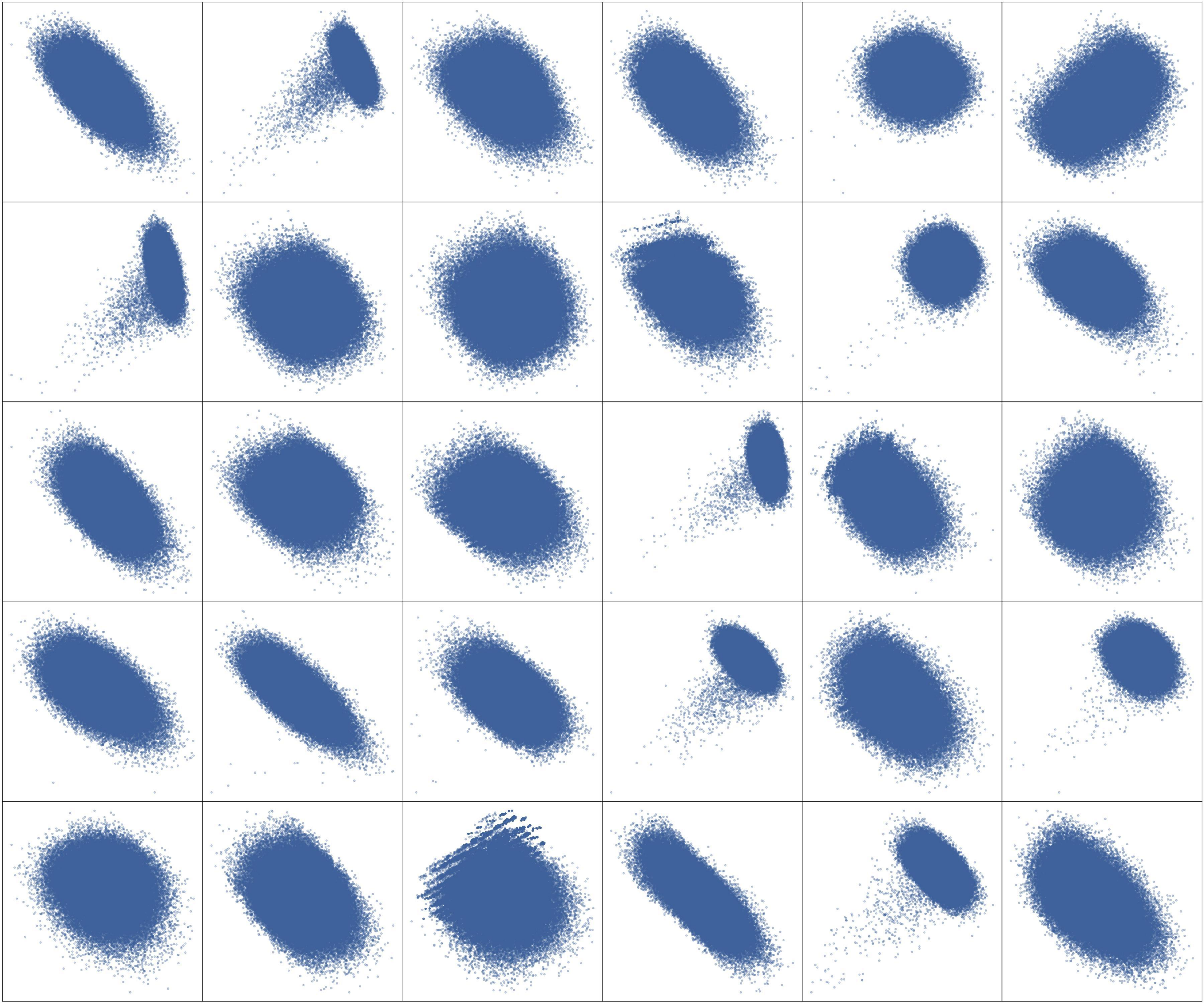}
    \label{fig:MKP_test_vis_100}}
    \caption{Objective-space distributions of all 120 MKP test instances. Each labeled panel contains 30 small plots, one per instance, showing the objective vectors of \(100{,}000\) randomly sampled binary solutions. (a) \(d=32\). (b) \(d=48\). (c) \(d=64\). (d) \(d=100\).}
    \label{fig:MKP_test_vis}
\end{figure*}

\begin{figure*}[htbp]
    \centering
    \includegraphics[width=0.70\textwidth]{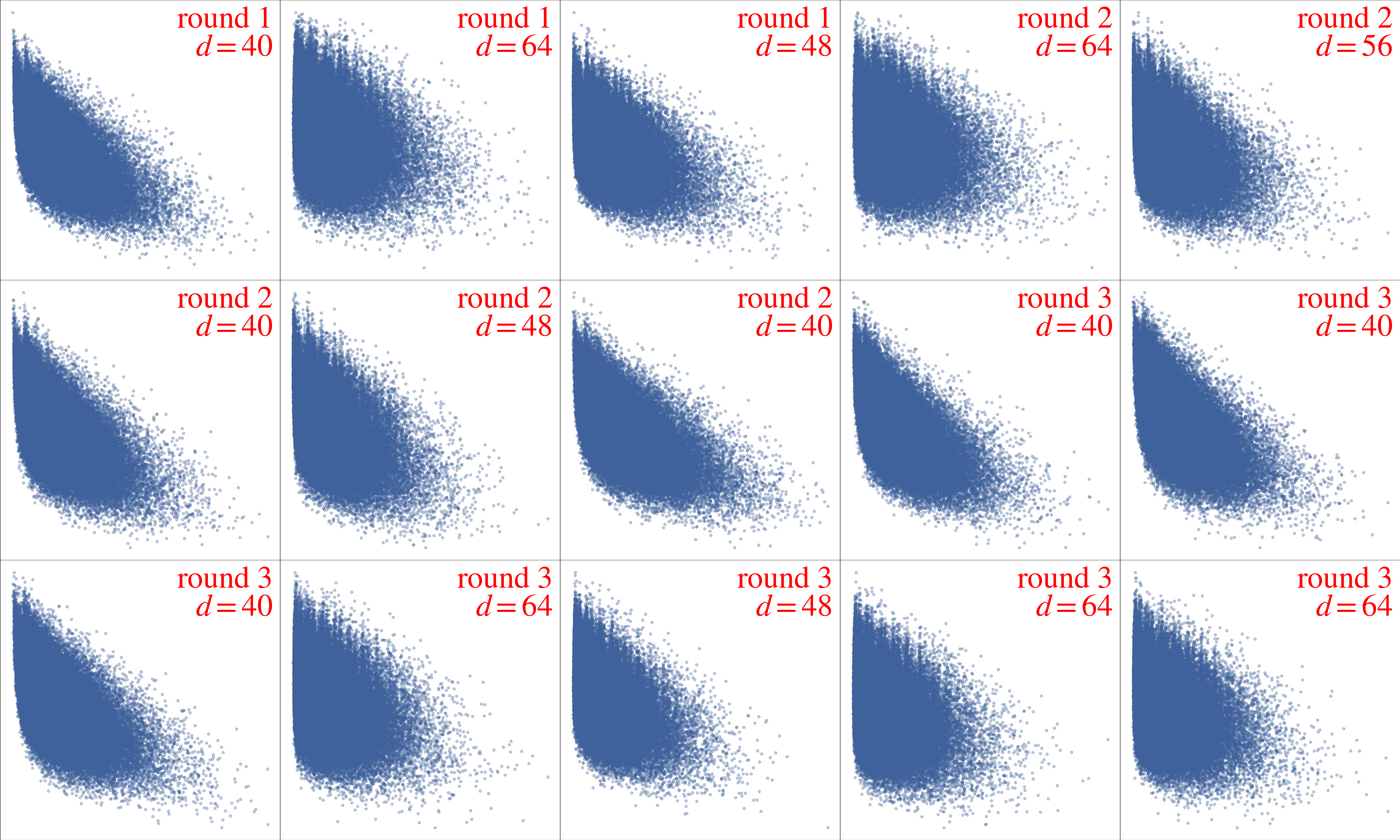}
    \caption{Objective-space distributions of the 15 UIR instances added to the MCCP training set during DACMO's co-evolution. Each plot corresponds to one instance and is labeled with its augmentation round and decision-variable dimension \(d\).}
    \label{fig:MCCP_augment}
    \par\medskip
    \subfloat[]{\includegraphics[width=0.40\textwidth]{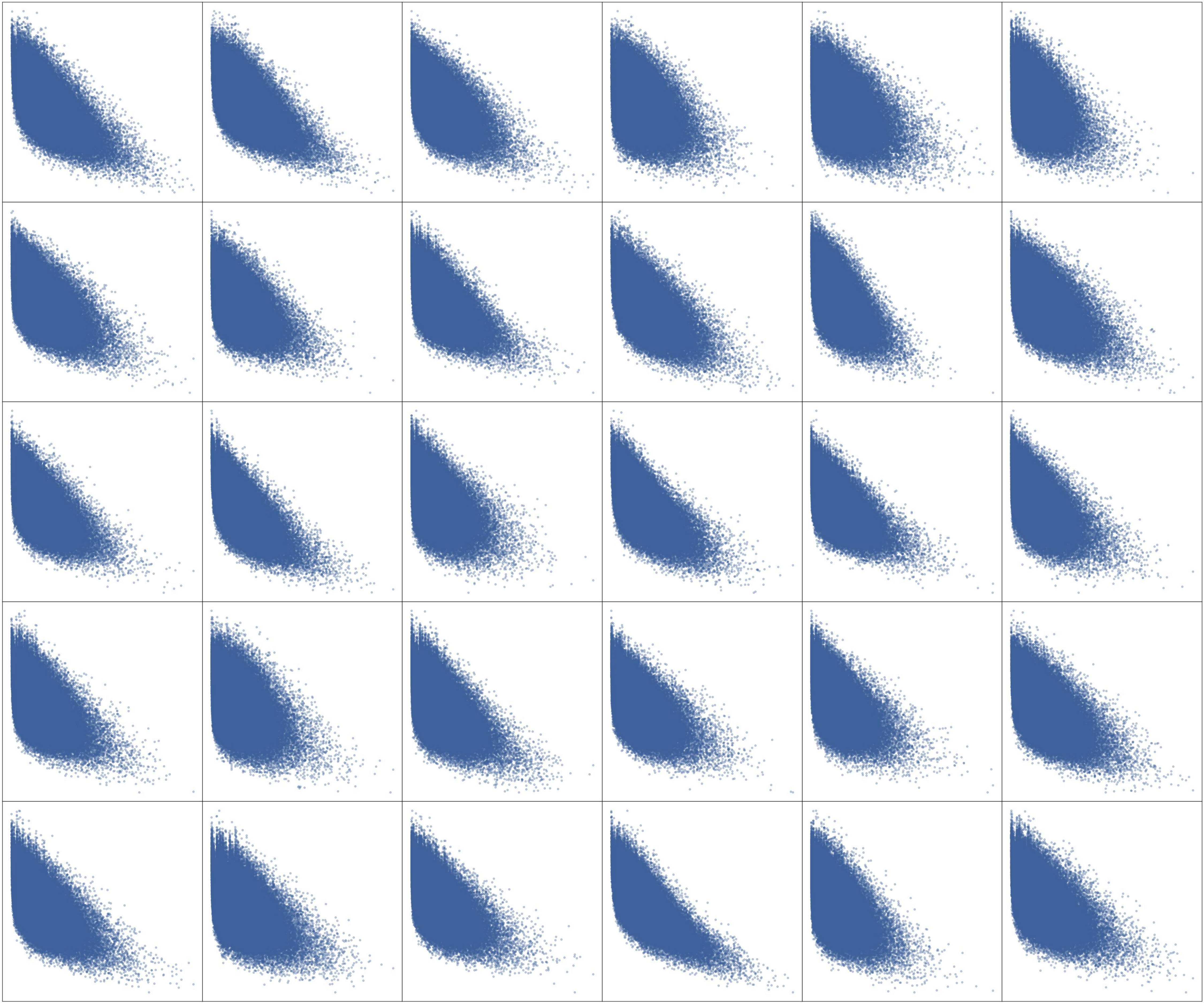}
    \label{fig:MCCP_test_vis_32}}
    \subfloat[]{\includegraphics[width=0.40\textwidth]{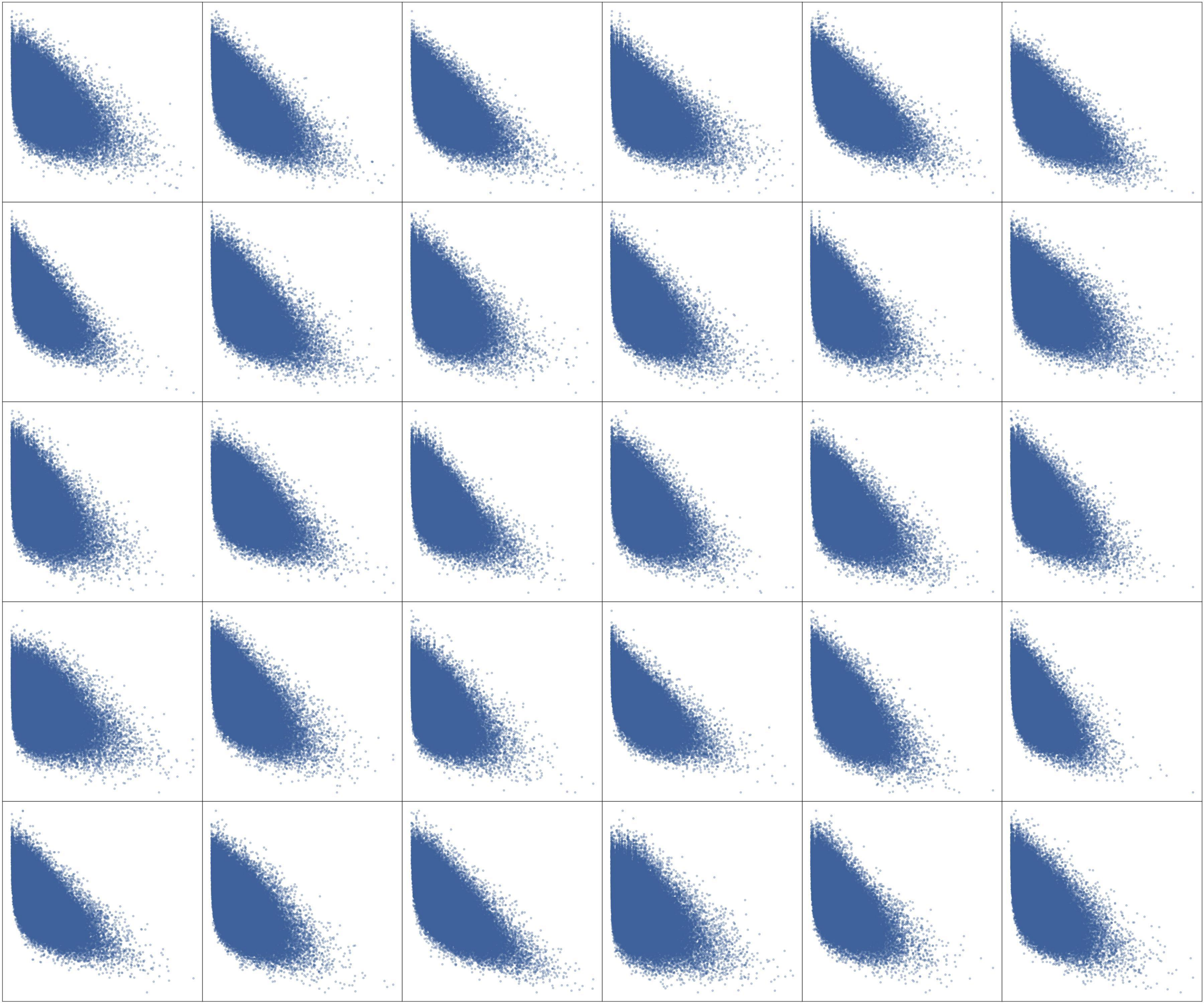}
    \label{fig:MCCP_test_vis_48}}\\
    \subfloat[]{\includegraphics[width=0.40\textwidth]{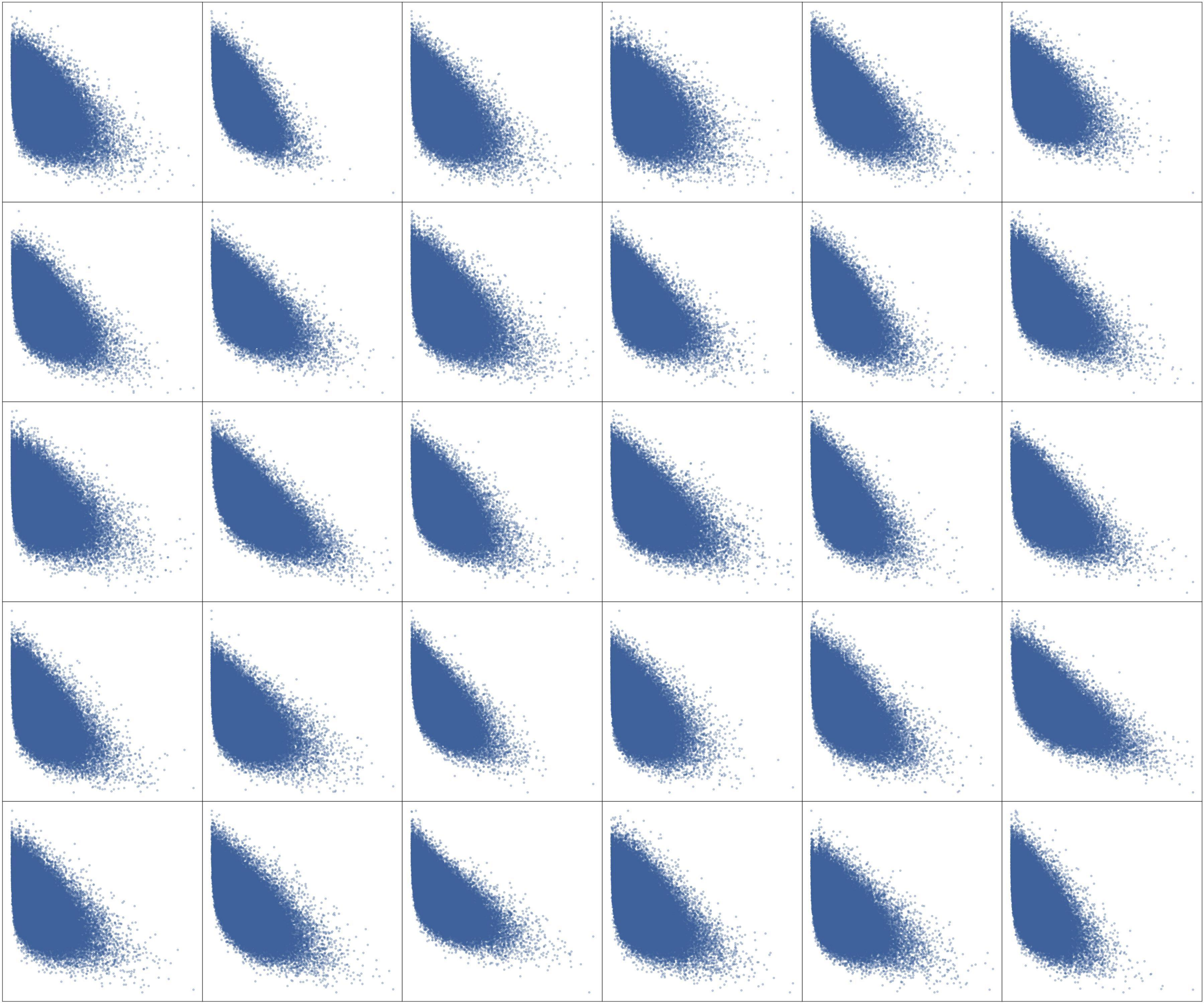}
    \label{fig:MCCP_test_vis_64}}
    \subfloat[]{\includegraphics[width=0.40\textwidth]{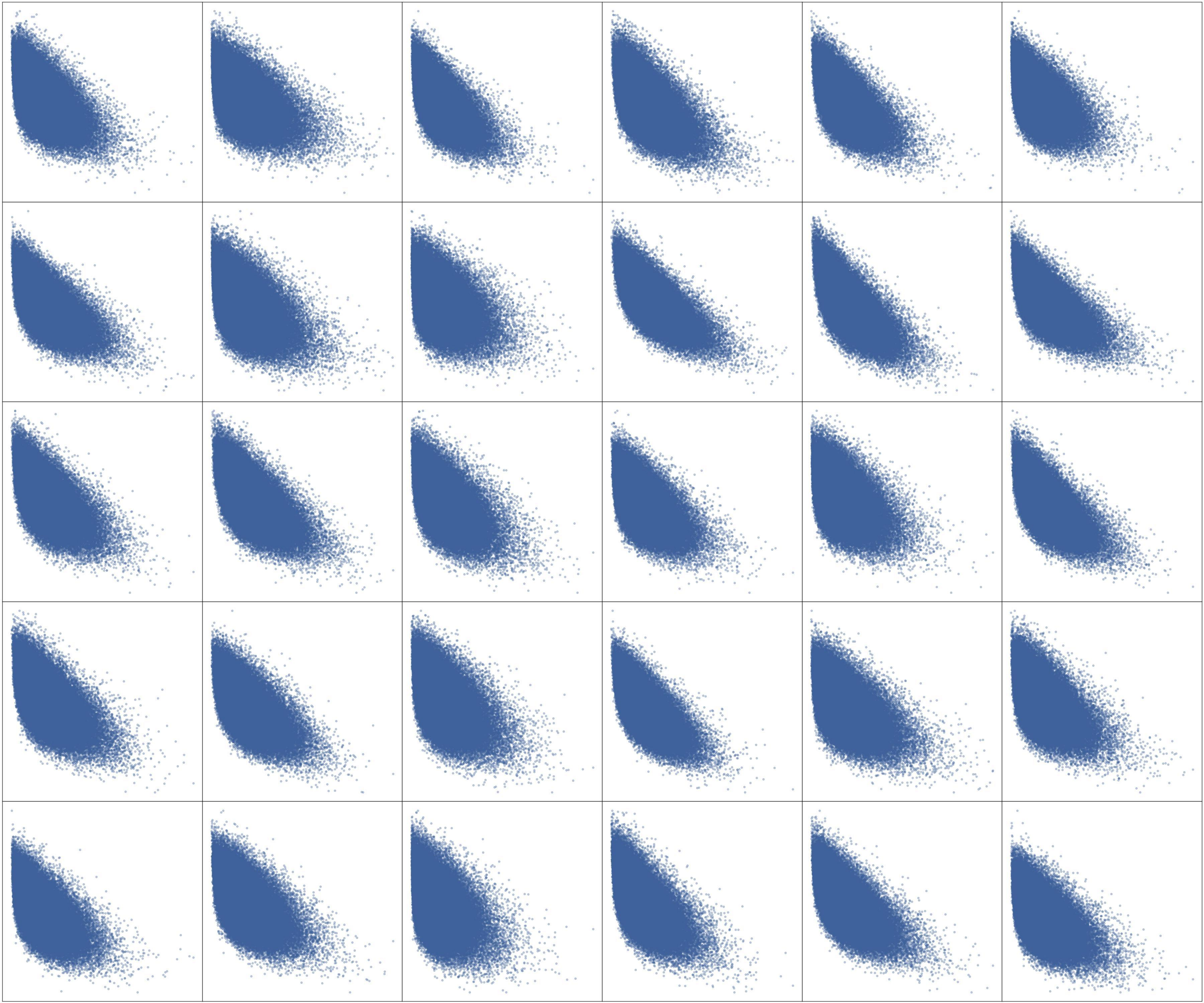}
    \label{fig:MCCP_test_vis_100}}
    \caption{Objective-space distributions of all 120 MCCP test instances. Each labeled panel contains 30 small plots, one per instance, showing the objective vectors of \(100{,}000\) randomly sampled binary solutions. (a) \(d=32\). (b) \(d=48\). (c) \(d=64\). (d) \(d=100\).}
    \label{fig:MCCP_test_vis}
\end{figure*}

\begin{figure*}[htbp]
    \centering
    \includegraphics[width=0.70\textwidth]{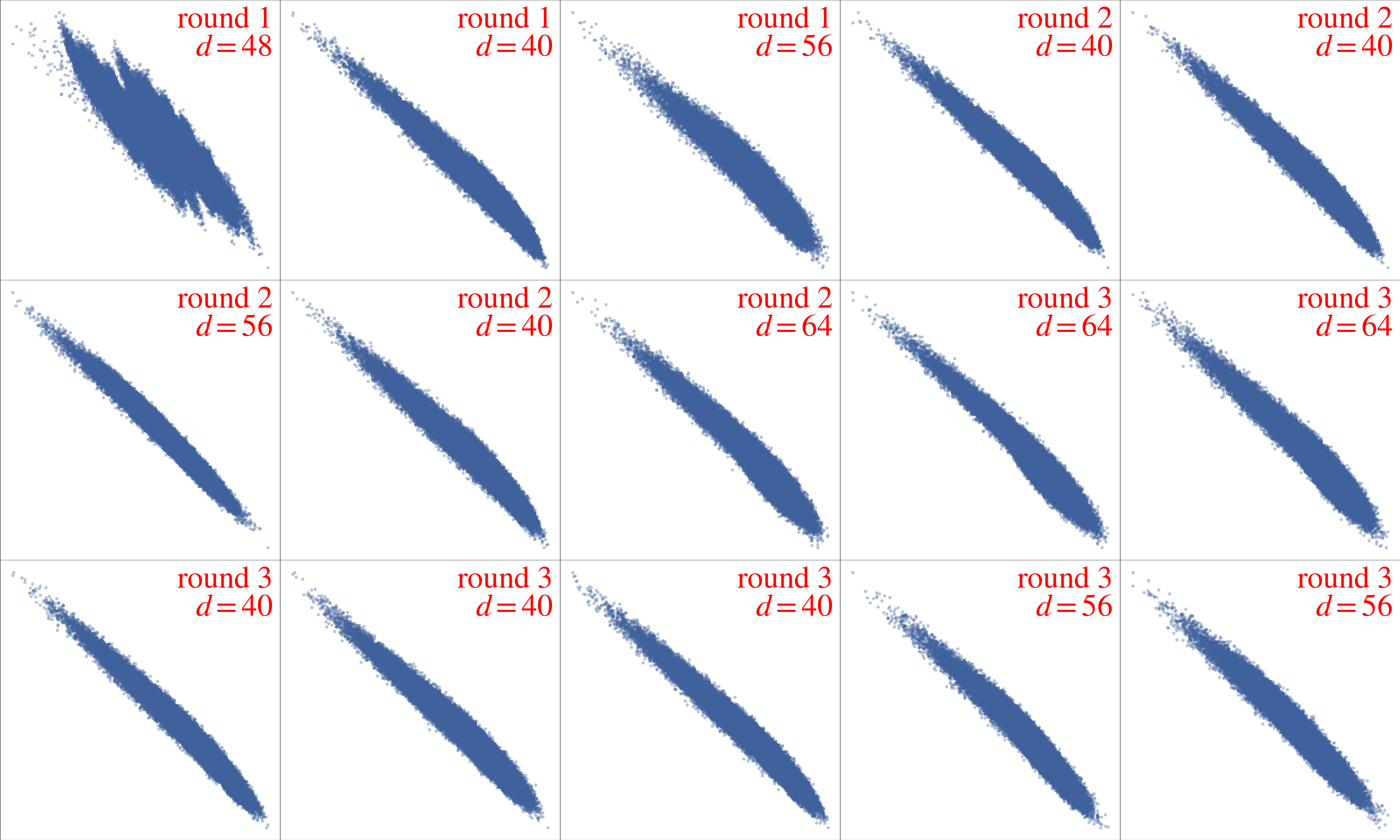}
    \caption{Objective-space distributions of the 15 UIR instances added to the MCIMP training set during DACMO's co-evolution. Each plot corresponds to one instance and is labeled with its augmentation round and decision-variable dimension \(d\).}
    \label{fig:MCIMP_augment}
    \par\medskip
    \subfloat[]{\includegraphics[width=0.40\textwidth]{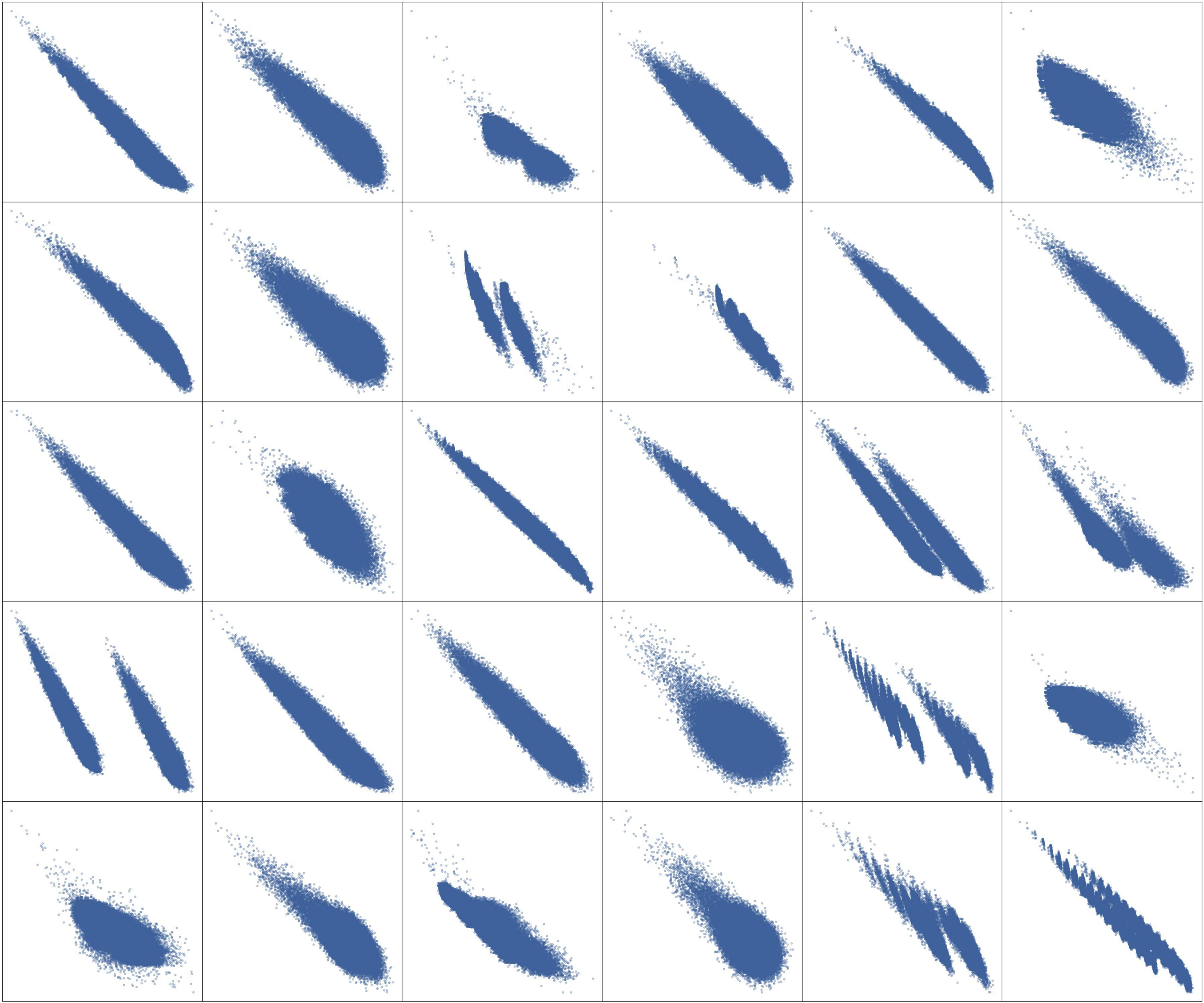}
    \label{fig:MCIMP_test_vis_32}}
    \subfloat[]{\includegraphics[width=0.40\textwidth]{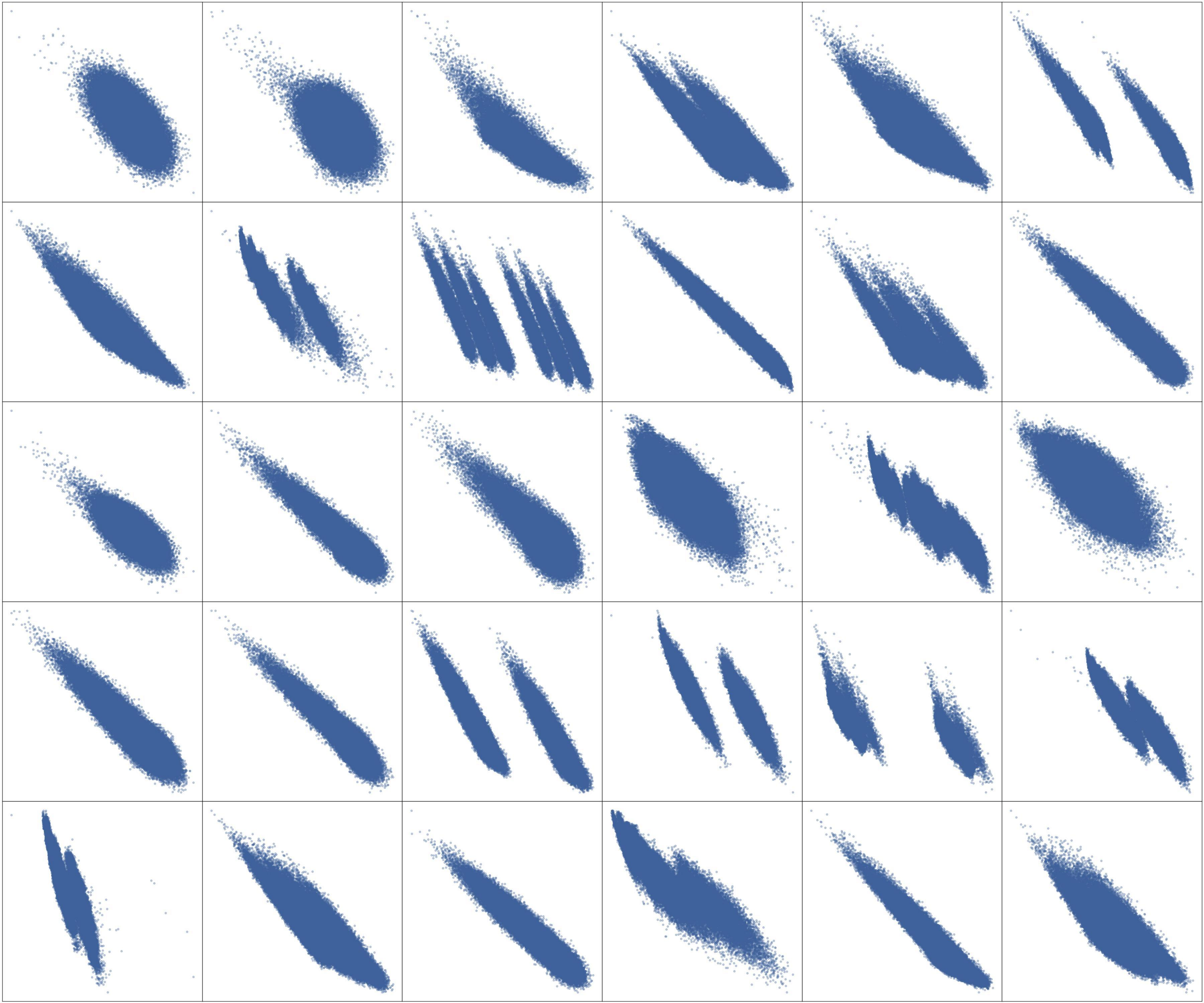}
    \label{fig:MCIMP_test_vis_48}}\\
    \subfloat[]{\includegraphics[width=0.40\textwidth]{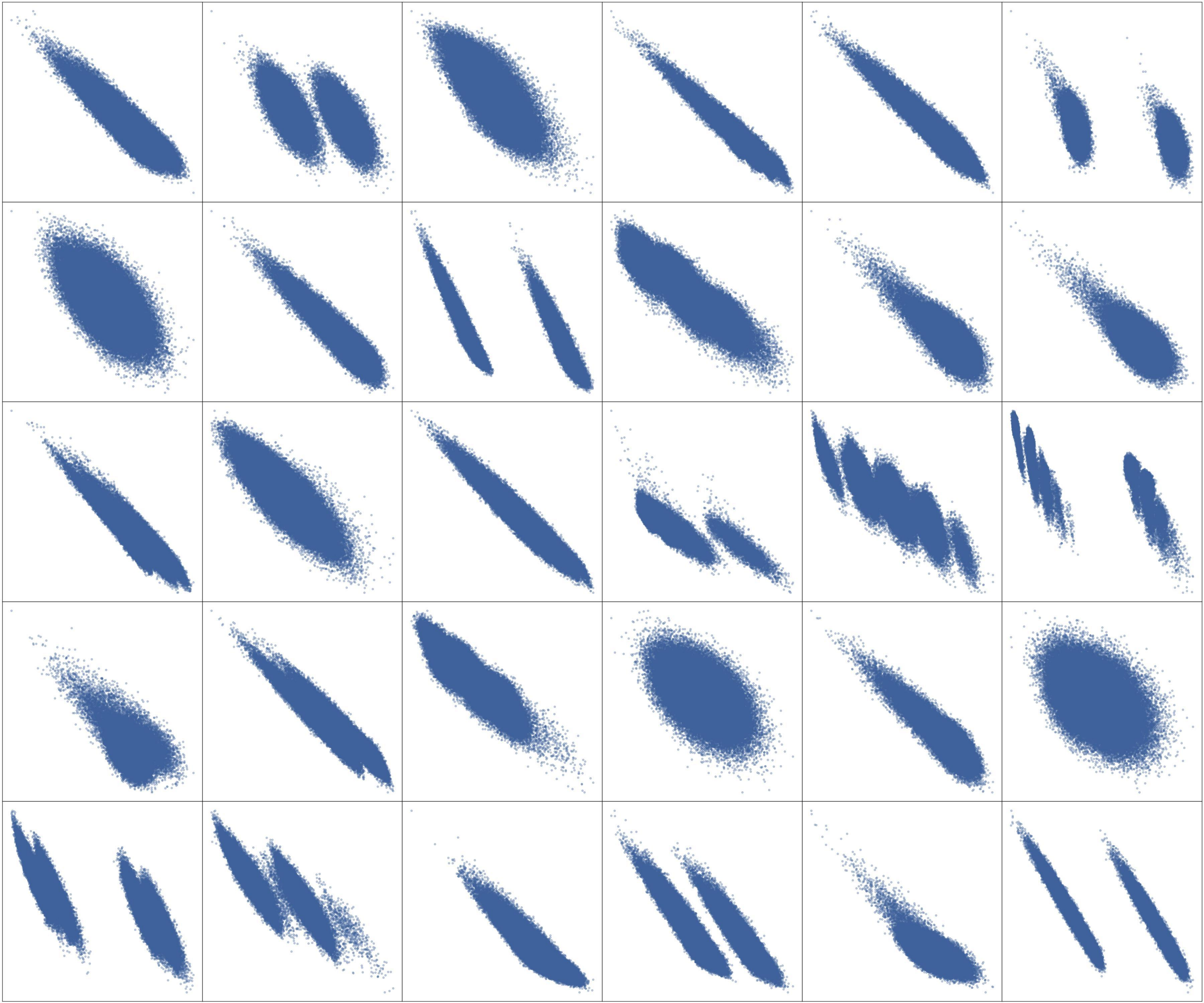}
    \label{fig:MCIMP_test_vis_64}}
    \subfloat[]{\includegraphics[width=0.40\textwidth]{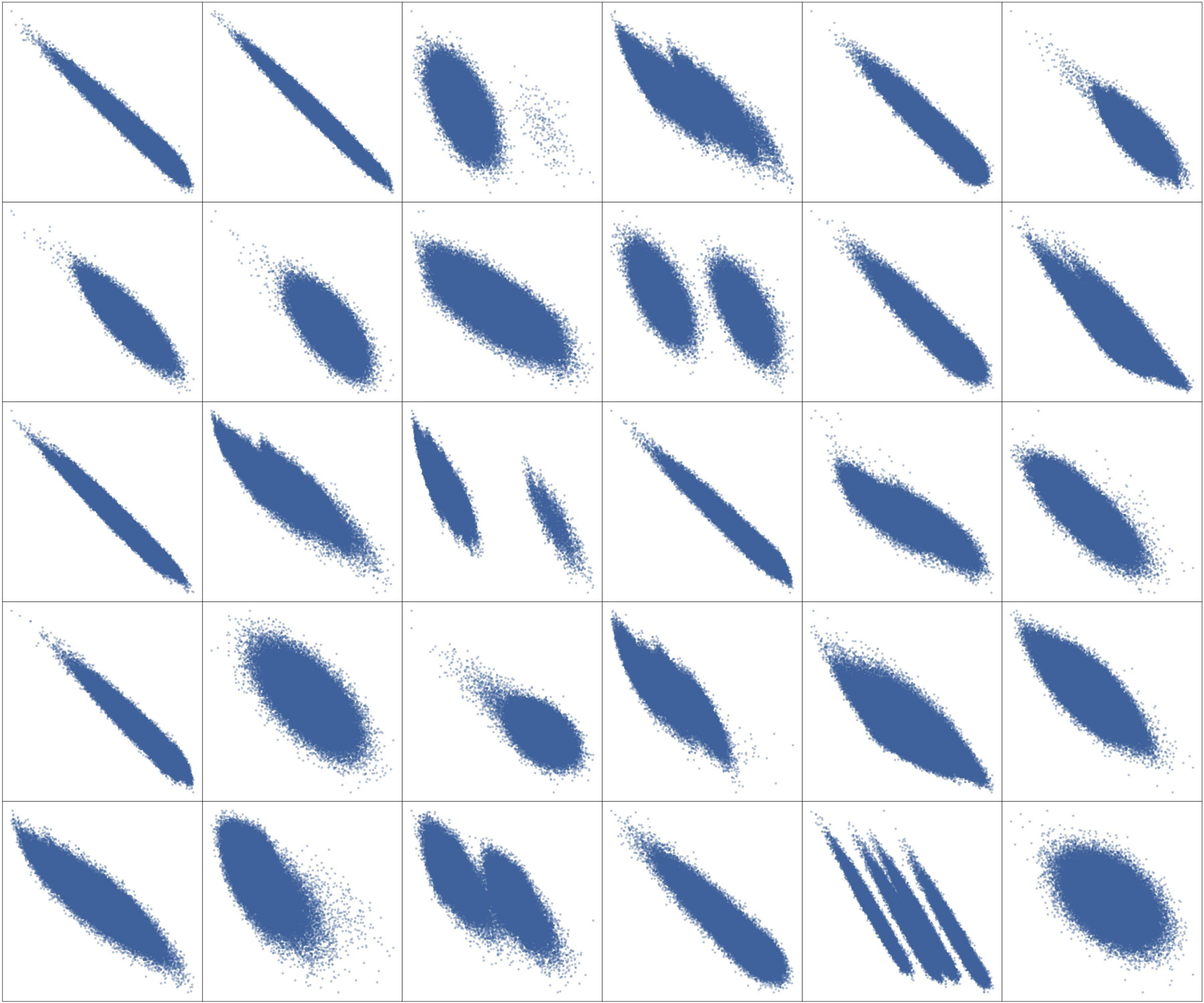}
    \label{fig:MCIMP_test_vis_100}}
    \caption{Objective-space distributions of all 120 MCIMP test instances. Each labeled panel contains 30 small plots, one per instance, showing the objective vectors of \(100{,}000\) randomly sampled binary solutions. (a) \(d=32\). (b) \(d=48\). (c) \(d=64\). (d) \(d=100\).}
    \label{fig:MCIMP_test_vis}
\end{figure*}

Figs.~\ref{fig:MKP_augment}--\ref{fig:MCIMP_test_vis} complement the visualization in Section~V-E of the main paper. For each of MKP, MCCP, and MCIMP, the 15 UIR instances added during DACMO's co-evolution are shown first, followed by the corresponding real test instances. Each augmented-instance plot is labeled with its augmentation round and number of decision variables. Each class has 30 test instances at each dimension of 32, 48, 64, and 100. Each small test-instance plot shows the objective vectors of \(100{,}000\) randomly sampled binary solutions, and the four labeled panels group these plots by dimension.

For MMMP, we quantify the similarity of all 15 augmented UIR instances to real instances using the objective function matching procedure in Section~V-E of the main paper. For each UIR instance, BRKGA searches for a matching real MMMP instance using \(10{,}000\) sampled binary solutions. The resulting match is evaluated on \(300{,}000\) independently sampled solutions, with per-objective min--max normalization recomputed for each instance on the validation set. Table~\ref{tab:mmmp_augment_similarity} reports the validation MSE aggregated by dimension.

\begin{table}[tbp]
    \centering
    \caption{Validation MSE between augmented UIR instances and their matched real MMMP instances. Means and population standard deviations are computed across the augmented instances of each dimension.}
    \label{tab:mmmp_augment_similarity}
    {
    \footnotesize
    \setlength{\aboverulesep}{0.05pt}
    \setlength{\belowrulesep}{0.05pt}
    \renewcommand{\arraystretch}{1.1}
    \setlength{\tabcolsep}{3pt}
    \begin{tabular*}{\columnwidth}{@{\extracolsep{\fill}}ccc@{}}
        \toprule
        Dim & \#Instances & Mean $\pm$ Std. \\
        \midrule
        32 & 1 & \(\text{6.8e-4}\pm\text{0}\) \\
        40 & 2 & \(\text{4.5e-3}\pm\text{4.5e-4}\) \\
        48 & 6 & \(\text{7.5e-4}\pm\text{1.6e-4}\) \\
        56 & 2 & \(\text{2.3e-3}\pm\text{2.0e-4}\) \\
        64 & 4 & \(\text{1.6e-3}\pm\text{3.3e-4}\) \\
        \bottomrule
    \end{tabular*}
    }
\end{table}

\end{document}